%% file: main.tex
\documentclass{article}

\PassOptionsToPackage{numbers, compress, sort}{natbib}
\usepackage[final]{neurips_2024}

\usepackage[utf8]{inputenc} 
\usepackage{colortbl}
\usepackage[T1]{fontenc}    
\usepackage{hyperref}       
\usepackage{url}            
\usepackage{booktabs}       
\usepackage{amsfonts}       
\usepackage{nicefrac}       
\usepackage{microtype}      
\usepackage{xcolor}         
\usepackage{graphicx} 
\usepackage{amsmath}
\usepackage{wrapfig}
\usepackage{multirow}
\usepackage{nicematrix}
\usepackage{uoftcolors}

\hypersetup{
  colorlinks=true,
  allcolors=uoftoceanblue
}

\title{GaussianCut: Interactive Segmentation \\via Graph Cut for 3D Gaussian Splatting}

\author{%
  Umangi Jain, Ashkan Mirzaei, and Igor Gilitschenski  \\
  University of Toronto\\
  \texttt{\{umangi, ashkan, gilitschenski\}@cs.toronto.edu} \\
}

\begin{document}

\maketitle
\begin{figure}[h]
    \centering
    \includegraphics[width=\linewidth]{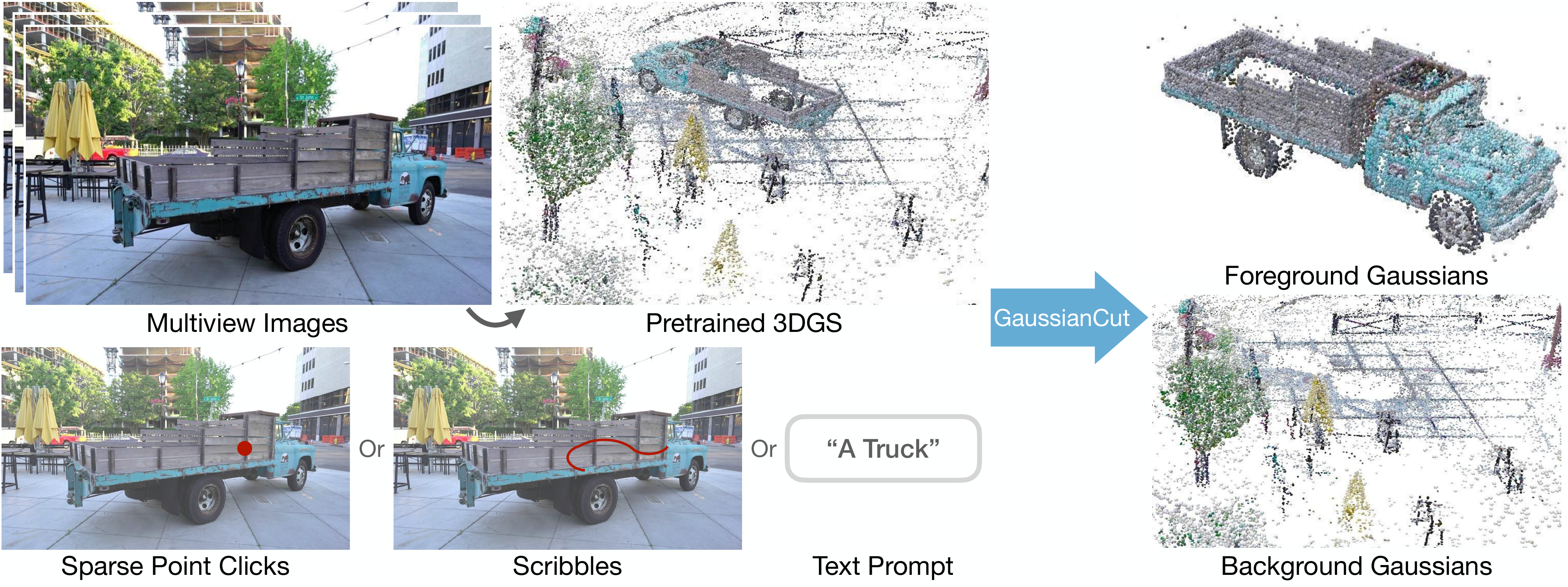}
    \caption{Our method, GaussianCut, enables interactive object(s) selection. Given an optimized 3D Gaussian Splatting model for a scene with user inputs (clicks, scribbles, or text) on any viewpoint, GaussianCut partitions the set of Gaussians as foreground and background. }
    \label{fig:teaser}
\end{figure}

\begin{abstract}
We introduce GaussianCut, a new method for interactive multiview segmentation of scenes represented as 3D Gaussians. Our approach allows for selecting the objects to be segmented by interacting with a single view. It accepts intuitive user input, such as point clicks, coarse scribbles, or text. Using 3D Gaussian Splatting (3DGS) as the underlying scene representation simplifies the extraction of objects of interest which are considered to be a subset of the scene's Gaussians. 
Our key idea is to represent the scene as a graph and use the graph-cut algorithm to minimize an energy function to effectively partition the Gaussians into foreground and background.
To achieve this, we construct a graph based on scene Gaussians and devise a segmentation-aligned energy function on the graph to combine user inputs with scene properties. 
To obtain an initial coarse segmentation, we leverage 2D image/video segmentation models and further refine these coarse estimates using our graph construction. Our empirical evaluations show the adaptability of GaussianCut across a diverse set of scenes. GaussianCut achieves competitive performance with state-of-the-art approaches for 3D segmentation without requiring any additional segmentation-aware training.

\end{abstract}

\section{Introduction}

Recent advances in 3D scene representation have enabled unprecedented quality in 3D view synthesis without requiring specialized equipment or an excessively high computational budget. 
Fully leveraging these advances requires tools for scene understanding and manipulation specifically designed to operate on such representations.
Object selection and segmentation often serve as a crucial first step in both scene understanding and editing tasks. 
While 2D image segmentation has been widely studied, developing analogous techniques for 3D remains challenging. One key challenge is accounting for the choice of underlying 3D scene representation in the segmentation method.


3D Gaussian Splatting (3DGS)~\cite{kerbl20233d} offers an explicit representation of a scene using a set of Gaussians, each characterized by its own properties. The nature of this representation motivates the idea that Gaussians corresponding to the segmented object and the background can be isolated separately.  Prior works in 3DGS segmentation involve augmenting each Gaussian with a low-dimensional feature, that is jointly optimized with the parameters of the Gaussians~\cite{ye2023gaussian,qin2023langsplat,cen2023segment}. This is supervised by 2D features, which provide semantic information that can be used for segmentation.  While this enables a 3D consistent segmentation, it significantly increases the fitting time and the already high memory footprint of the method. Thus, enabling 3DGS segmentation without modifying the optimization process is an important research challenge. 

We address this challenge by proposing GaussianCut, a novel method for selecting and segmenting objects of interest in 3D Gaussian scenes. Our work taps directly into the representation created by 3DGS and maps each Gaussian to either the foreground or background.
The proposed process mirrors the interactive nature of 2D segmentation tools, where users can engage through clicks, prompts, or scribbles. We require such user input on a single image and perform the object selection process in two steps. 
First, we obtain dense multiview segmentation masks from the user inputs using a video segmentation model. Subsequently, we construct a weighted graph, where each node represents a Gaussian. Graph cut then partitions the graph into two disjoint subsets by minimizing an energy function, which quantifies the cost of cutting the edges connecting the subsets.  This approach effectively segments the selected foreground object from the background by using the energy function as a measure of dissimilarity between the nodes. An overview of the process is provided in Figure~\ref{fig:teaser}.


Our main contribution is a novel approach for segmentation in scenes obtained from 3DGS. Its main technical novelties are twofold:
1) we propose a method for graph construction from a 3DGS model that utilizes the properties of the corresponding Gaussians to obtain edge weights, and 2) based on this graph, we propose and minimize an energy function (Equation~\ref{eq:energy_function}) that combines the user inputs with the inherent representation of the scene.
Our experimental evaluations show that GaussianCut obtains high-fidelity segmentation outperforming previous segmentation baselines.

\section{Related Work}
2D image segmentation is a long studied problem in computer vision~\cite{garcia2017review,peng2013survey,zaitoun2015survey}. Recently, models like Segment Anything~\cite{kirillov2023segment} and SEEM~\cite{zou2024segment}, have revolutionized 2D segmentation by employing interactive segmentation. A range of methods have also been developed for 3D segmentation, each tailored to different forms of representation, including voxels~\cite{milletari2016v,cciccek20163d}, point clouds~\cite{qi2017pointnet,qi2017pointnet++}, meshes~\cite{tang2021dense,zhou2023serf}, and neural representations~\cite{liu2023instance,vora2021nesf,cen2024segment,stelzner2021decomposing}. The impressive capabilities of Neural Radiance Fields (NeRFs)~\cite{mildenhall2021nerf} in capturing scene information implicitly have inspired numerous studies to explore 3D segmentation for NeRFs. Recent works have also explored segmentation with Gaussians as the choice for scene representation~\cite{ye2023gaussian,zhou2023feature,cen2023segment,hu2024semantic,qin2023langsplat}.


\textbf{Training 3D segmentation with 2D masks/features: } In addition to the wide adaptation of foundational models for 2D images~\cite{zhang2023survey}, they are also used extensively by 3D editing and segmentation models. SAM has been used as an initial mask to facilitate 3D segmentation~\cite{cen2023segment, liu2023sanerf, cen2024segment} and also for distillation into NeRF~\cite{zhang2024open} and 3DGS models~\cite{zhou2023feature}. Semantic-NeRF~\cite{zhi2021place} proposed 2D label propagation to incorporate semantics within NeRF so it can produce 3D consistent masks. MVSeg~\cite{mirzaei2023spin} propagates a 2D mask to different views using video segmentation techniques. 
ISRF~\cite{goel2023interactive} distills semantic features into the 3D scene of voxelized radiance fields. Nearest neighbor feature matching then identifies high-confidence seed regions. 2D features have also been used for facilitating the grouping of Gaussians~\cite{ye2023gaussian} and for hierarchical semantics using language in 3DGS~\cite{qin2023langsplat}. Distilled Feature Fields (DFF)~\cite{kobayashi2022decomposing} and Neural Feature Fusion Fields~\cite{tschernezki2022neural} distill 2D image embeddings from LSeg~\cite{liu2022open} and DINO~\cite{caron2021emerging} to enable segmentation and editing. SA3D~\cite{cen2024segment} uses SAM iteratively to get 2D segments and then uses depth information to project these segments into 3D mesh grids.   
SANeRF-HQ~\cite{liu2023sanerf} aggregates 2D masks in 3D space to enable segmentation with NeRFs.

\textbf{Segmentation in 3D Gaussian Splatting: } Gaussian Grouping~\cite{ye2023gaussian}, SAGA~\cite{cen2023segment}, LangSplat~\cite{qin2023langsplat}, CoSSegGaussians~\cite{dou2024cossegGaussians} and Feature 3DGS~\cite{zhou2023feature} require optimizing a 3DGS model with an additional identity or feature per Gaussian, which is usually supervised by 2D image features. These semantic features allow segmentation through user interaction. Gaussian Grouping and LangSplat also allow for textual prompts to segment objects supported through multimodal models like CLIP or grounding-DINO~\cite{liu2023grounding}.  Feature-based methods alter the fitting process of 3DGS by adding additional attributes for each Gaussian and it facilitates learning features for everything in the scene. While useful, this limits the flexibility of interactivity with a single object. Our method is more flexible in choosing specific object(s) as we generate the 2D masks after the user interaction. Adding additional parameters also increases the fitting time for 3DGS. Moreover, such methods often rely on video segmentation models as they require 2D features from all training viewpoints. In contrast, we can operate on an arbitrary number of 2D masks, including just a single mask.

\textbf{Graph cut for 3D segmentation: } Boykov and Jolly~\cite{boykov2001interactive} introduced a novel global energy function for interactive image segmentation using graph cut~\cite{ford2015flows,goldberg1988new}. Several follow-up works improved image segmentation using graph cut by designing better energy function~\cite{felzenszwalb2004efficient}, efficient optimization~\cite{grady2006random,boykov2001fast}, and reduced user input requirements~\cite{tu2008auto}. Adapting energy minimization methods for 3D volumes has been difficult, requiring several modifications~\cite{grady2006random} to manage the higher memory demands. NVOS~\cite{ren2022neural} trains a special multi-layer perceptron (MLP) to predict voxels in the foreground and background and applies graph cut on voxels as post-processing. However, training the MLP requires additional training and memory consumption. Guo \textit{et al.}~\cite{guo2023sam} propose 3D instance segmentation of 3D point clouds using graph cut. It involves constructing a superpoint graph and training a separate graph neural network for predicting the edge weights. Unlike their work, our method is a post hoc application and does not require any additional training. Our graph construction and edge weights have also been tailored specifically for 3D Gaussian Splatting. 

Concurrent with our work, Segment Anything in 3D Gaussians (SAGD)~\cite{hu2024semantic}, also performs interactive segmentation using 3D Gaussian Splatting without requiring any segmentation-aware training. However, their focus is primarily on refining object boundaries by decomposing boundary Gaussians, whereas we propose a graph cut based approach for interactive segmentation.




\section{Method}

\subsection{Preliminaries}

\textbf{3D Gaussian Splatting (3DGS)}~\cite{kerbl20233d} is a technique for creating a 3D representation of scenes based on a set of Gaussian ellipsoids $\mathcal{G}$. 
3DGS facilitates real-time rendering and provides high-quality reconstruction. In this representation, each 3D Gaussian is characterized by a set of optimizable parameters that include 3D position  $\boldsymbol{\mu} \in \mathbb{R}^3$, spherical harmonics (SH) coefficients (for color) $\boldsymbol{\beta} \in \mathbb{R}^{3(d^{2}+1)}$ ($d$ is the degree of spherical harmonics), scale $\mathbf{s} \in \mathbb{R}^3$, rotation $\mathbf{r} \in \mathbb{R}^4$, and opacity $\sigma \in \mathbb{R}$. The optimization process involves iteratively rendering scenes and comparing the rendered images against the training views, interleaved with adaptive density control that handles the creation and deletion of the number of Gaussians. The differentiable rendering pipeline in 3DGS uses tile-based rasterization following~\cite{lassner2021pulsar} to ensure real-time rendering. 3DGS performs anisotropic splatting by depth sorting the Gaussians and $\alpha$-blending them to project in 2D. The set of differentiable parameters for $\mathcal{G}$, $\mathcal{D} := \{\boldsymbol{\mu}_i, \boldsymbol{\beta}_i, \mathbf{s}_i, \mathbf{r}_i\, \sigma_i\}_{i=1}^{|\mathcal{G}|}$, are optimized from a set of posed images. 
\noindent\textbf{Graph cut} is an algorithm that partitions the vertices $\mathcal{V}$ of a graph $\mathbf{G}$ with edges $\mathcal{E}$ weighted by $\{ w_e \}_{e \in \mathcal{E}}$ into two disjoint, non-empty sets such that the sum of the weights of the edges between the two sets is minimized. This minimum-cost partitioning is known as the \textit{minimum cut}.
 In applications such as image segmentation, the graph cut framework is adapted by defining an energy function, which includes unary terms representing the cost of assigning a node to a set based on individual properties, and pairwise terms that incorporate the cost of assigning neighboring nodes to different sets. The objective of the minimization is to find a cut that optimizes the overall energy, balancing individual preferences and neighborhood interactions. An efficient way for computing this minimum cut in a graph is the Boykov-Kolmogorov algorithm~\cite{boykov2004experimental}.


\subsection{Overview}
\begin{figure}[t]
    \centering
    \includegraphics[width=\linewidth]{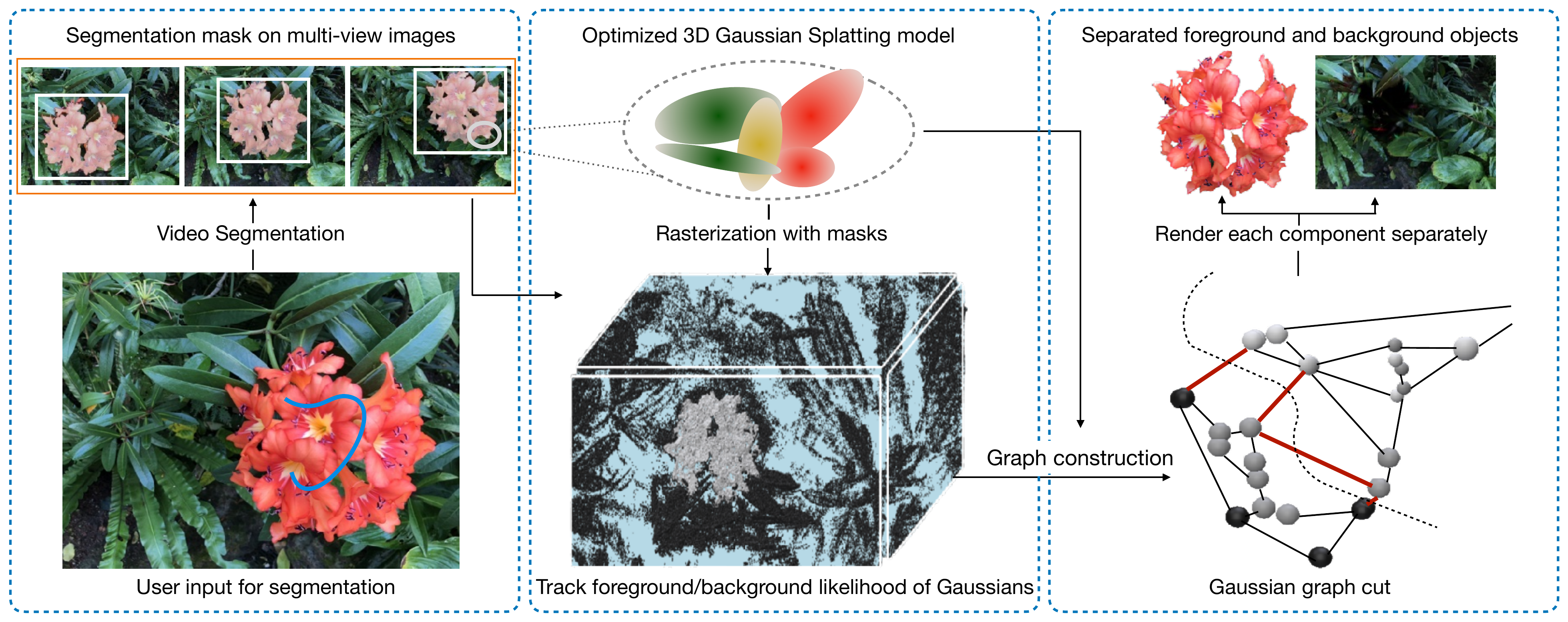}
    \caption{Overall pipeline of GaussianCut. User input from any viewpoint is passed to a video segmentation model to produce multi-view masks. We rasterize every view and track the contribution of each Gaussian to masked and unmasked pixels. Then, Gaussians are formulated as nodes in an undirected graph and we adapt graph cut to partition the graph. The red edges in the graph highlight the set of edges graph cut removes for partitioning the graph.}
    \label{fig:overview}
\end{figure}

%
%
Given a set of posed RGB images $\mathcal{I} = \{\mathbf{I}_i\}_{i=1}^{k}$ and an optimized reconstruction of the scene using a set of Gaussians $\mathcal{G}$, 
we define the task of interactive 3D segmentation as follows: given a user input (point clicks, scribbles, or text) on any image $\mathbf{I}_0 \in \mathcal{I}$, the objective is to partition the set of Gaussians in two non-empty disjoint sets $\mathcal{S}$ and $\mathcal{T}$ such that $\mathcal{S}$ represents the set of Gaussians representing 
the object(s) of interest and $\mathcal{T}$ represents the given 3D scene without these object(s). The extracted subset of Gaussians $\mathcal{S}$ can be rendered from any viewpoint to effectively cutout the 3D representation of the foreground without retraining. Innately, the other set of Gaussians  $\mathcal{T}$ can be rendered to remove the foreground object(s). Figure~\ref{fig:overview} shows an overview of our pipeline.


In order to densify the object selection information provided by the user, we use an off-the-shelf video segmentation model to obtain dense segmentation masks from multiple views (discussed in section~\ref{sec:setup}). For transferring the segmentation masks to the 3DGS representation, we trace the 3D Gaussians that map to the selected region in the masks (discussed in section~\ref{met:trace}). However, the masks used for propagation are not 3D consistent (as the underlying image segmentation model is 3D-unware). Moreover, the errors from 2D masks can propagate in the traced Gaussians and thereby provide a noisy 3D segmentation. To achieve a precise set of foreground Gaussians, we formulate the set of Gaussians $\mathcal{G}$ as nodes in a graph network (discussed in section~\ref{met:graph}) and leverage graph cut to split the nodes into two sets: the foreground $\mathcal{S}$ and the background $\mathcal{T}$. 





\subsection{Mapping user inputs to Gaussians}\label{met:trace}
We first feed the sparse single-view annotations by the user (e.g., point clicks) to a multiview/video segmentation model to obtain coarse segmentation masks across multiple training views. 
We then propagate the information from the 2D masks onto the set of Gaussians. For an already optimized 3DGS model of a scene, $\mathcal{G}$, we obtain $n$ masks $\mathcal{M} := \{\mathbf{M}^j \}_{j=1}^n$ from a video segmentation model
corresponding to any $n$ viewpoints $\mathcal{I} := \{\mathbf{I}^j \}_{j=1}^n$. Here, $\mathbf{M}^j$ indicates the set of foreground pixels in the viewpoint $\mathbf{I}^j$. For each Gaussian $g \in \mathcal{G}$, we maintain a weight, $w_g$, that indicates the likelihood of the Gaussian belonging to the foreground. To obtain the likelihood term $w_{g}^j$ pertaining to mask $j$ for Gaussian $g$, we unproject the posed image $\mathbf{I}^j$ back to the Gaussians using inverse rendering and utilizing the mask information,
\begin{align}
w_g^j = \frac{\sum_{\textbf{p} \in \mathbf{M}^j } \sigma_g(\textbf{p})T_g^j(\textbf{p}) }{\sum_{\textbf{p} \in {\mathbf{I}}^j} \sigma_g(\textbf{p})T_g^j(\textbf{p})},
\end{align}
where $\sigma_g(\textbf{p})$ and $T_g^j(\textbf{p})$ denote the opacity and transmittance from pixel $\textbf{p}$ for Gaussian $g$. Combining over all the $n$ masks, 
\begin{align}
w_g = \frac{\sum_{j}\sum_{\textbf{p} \in \mathbf{M}^j } \sigma_g(\textbf{p})T_g^j(\textbf{p})}{\sum_j\sum_{\textbf{p} \in {\mathbf{I}}^j} \sigma_g(\textbf{p})T_g^j(\textbf{p}) }.
\end{align}

This likelihood, $w_g$, captures the weighted ratio of the contribution of Gaussian $g$ to the masked pixels relative to the total number of pixels influenced by it.
%
The complementary value of $w_g$, $1-w_g$ provides the likelihood of Gaussian $g$ contributing to the background. The value of $w_g$ is updated using $n$ 2D segmentation masks during rasterization. Since the rasterization of 3DGS is remarkably fast, each pass typically takes less than a second to update. GaussianEditor~\cite{chen2024gaussianeditor} also learns an additional tracing parameter but unlike our approach, it maps Gaussians to different semantic classes.

Having the likelihoods $w_g$, a naive approach to extract the 3D representation of the foreground is to threshold the Gaussians and prune those with values below a certain threshold $\tau$.
We denote this approach as ``coarse splatting''. 
Figure~\ref{fig:bonsai_qual} demonstrates coarse splatting renders for a \textit{plant} in the 360-garden scene~\cite{barron2022mip}. Note that the renderings produced by coarse splatting are not accurate, particularly around the edges. This is due to two main reasons: 
1) the 2D segmentation models are 3D inconsistent and can be imperfect, leading to artifacts in the final Gaussian cutout, and 
2) lifting 2D masks to 3D can introduce its own artifacts.

\subsection{Gaussian graph construction}\label{met:graph}
After rasterization with the masks, each Gaussian $g \in \mathcal{G}$ in the 3DGS representation is characterized with parameters $\mathcal{D}_g := \{\boldsymbol{\mu}_g, \boldsymbol{\beta}_g, \mathbf{s}_g, \mathbf{r}_g\, \sigma_g, w_g\}$, where $w_g$ captures the user requirement and the other parameters encapsulate the inherent properties of the scene. To fuse the two sources of information and obtain a precise set of foreground Gaussians, we formulate the optimized 3DGS model as an undirected weighted graph $\mathbf{G}:=(\mathcal{G},\mathcal{E})$, where each Gaussian in $\mathcal{G}$ is a node and $\mathcal{E}$ represents the set of edges connecting spatially adjacent Gaussians. We define the neighborhood $\mathcal{N} \subseteq \mathcal{G} \times \mathcal{G}$ of a node (Gaussian) as its $k$-nearest  Gaussians in terms of their 3D position. The intuition behind constructing the edges is that Gaussians that map to the same object would be close spatially. 

Gaussian graph cut partitions the Gaussians  $\mathcal{G}$ into two disjoint and non-empty sets $\mathcal{S} \subset \mathcal{G}$ and $\mathcal{T} \subset \mathcal{G}$, that represent the foreground and background Gaussians, respectively. Our objective is to infer the
foreground/background label $y_g \in \{0, 1\}$ of each Gaussian $g$. Let the unary term $\mathbf{\phi}_g(\cdot, \cdot)$ represent the likelihood of node $g$ being part of foreground or background and the pairwise term $\mathbf{\psi}_{g, g'}(\cdot, \cdot)$ reflect the edge connection between node $g$ and $g'$. To obtain the label for each Gaussian $g$, graph cut minimizes the aggregate of both unary and pairwise terms given by:
\begin{align}
    E = \sum_{g \in \mathcal{G}} \mathbf{\phi}_g(\mathcal{D}_g, y_g) + \lambda \sum_{\substack{g,g' \in \mathcal{N}}} \mathbf{\psi}_{g,g'}(\mathcal{D}_g, \mathcal{D}_{g'}), \label{eq:energy_function}
\end{align}
where $\lambda$ provides a trade-off between the two terms.

 \textbf{Neighboring pairwise weights }\textit{(n-links)}: The pairwise term models the correlation between neighboring nodes. The neighbors for a node are based on its spatial proximity to other nodes. The edge weight between each pair of neighbors is a combination of its spatial distance and color similarity. While segments of an object can have multiple colors, and they often do, neighboring nodes with dissimilar colors can still be identified and grouped based on their spatial proximity. This ensures that parts of an object, despite varying in color, can be linked if they are close in space. For color similarity, we only use the zero-degree spherical harmonic to capture the ambient colors without any angular dependence. The correlation between the neighboring nodes is formulated as
\begin{align}
    \mathbf{\psi}_{g,g'}(\mathcal{D}_g, \mathcal{D}_{g'}) = \mathbf{f}(\boldsymbol{\mu}_g, \boldsymbol{\mu}_{g'}) + \lambda_n\mathbf{f}(\boldsymbol{\beta}_g, \boldsymbol{\beta}_g') ,
    \label{eq:nlink}
\end{align}
where $\lambda_n$ is a hyperparameter balancing the contribution of position and color similarity, and the function $\mathbf{f}$ estimates similarity as $\mathbf{f}(\mathbf{x}, \mathbf{y}) = \exp(-\gamma \|\mathbf{x} - \mathbf{y}\|_2^2)$  ($\gamma$ is a positive scalar).

 \textbf{Unary weights }\textit{(t-links)}: We designate two terminal nodes for the graph cut algorithm, the source and the sink node. These terminals represent the foreground (source) and the background (sink) in segmentation tasks. $t$-links connect all the nodes to both the terminal nodes and the edge weight for these links represents the pull of that node towards each terminal node. We assign the edge weights connecting each non-terminal node to the source node to reflect its likelihood of belonging to the foreground set $\mathcal{S}$ (and as belonging to the background set $\mathcal{T}$ for edges connecting to the sink node).
 
Gaussian tracking, from section~\ref{met:trace}, provides the connection of each Gaussian $g$ to the source and sink terminal nodes, using $w_{g}$ and $1 - w_g$, respectively. However, these weights can be noisy estimates. Therefore, we introduce an additional term to the edge weights that captures the similarity of node $g$ to the other nodes that are well-connected to the terminal nodes. To do so, we identify high-confidence nodes for both the source and the sink terminals. A Gaussian $g$ is considered as a high-confidence node for the source terminal if $w_g \approx 1 $ and for a sink terminal if $w_g \approx 0$. Since computing the similarity of a node to all the high-confidence nodes is computationally expensive, we cluster all the high-confidence nodes (denoted as $\mathcal{F}$ and $\mathcal{B}$ for the source and sink, respectively) based on their position. For each node 
$g$, we then determine the closest cluster by finding  $g_f = \mathrm{argmin}_{g' \in \mathcal{F}}\mathbf{f}(\boldsymbol{\mu}_g, \boldsymbol{\mu}_{g'})$ for the source, and similarly $g_b$ for the sink. 
Consequently, the unary term based on the user input is,
\begin{align}
\mathbf{\phi}_g(\mathcal{D}_g, y_g) = 
    \begin{cases}
    w_g + \lambda_u {\psi}_{g,g_f}(\mathcal{D}_g, \mathcal{D}_{g_f}) & \text{if } y_g = 1, \\
    1-w_g + \lambda_u{\psi}_{g,g_b}(\mathcal{D}_g, \mathcal{D}_{g_b}) & \text{if } y_g = 0. 
    \end{cases} 
    \label{eq:tlink}
\end{align}

We minimize the objective $E$ in Equation~\ref{eq:energy_function} to partition the set of nodes $\mathcal{G}$ as $\mathcal{S}$ (foreground Gaussians) and $\mathcal{T}$ (background Gaussians). To render the foreground object from any viewpoint, we simply render the Gaussian collected in $\mathcal{S}$ with $\mathcal{T}$ as background.






\section{Experimental Setup}\label{sec:setup}
\begin{figure}[t]
    \centering
    \includegraphics[width=\linewidth]{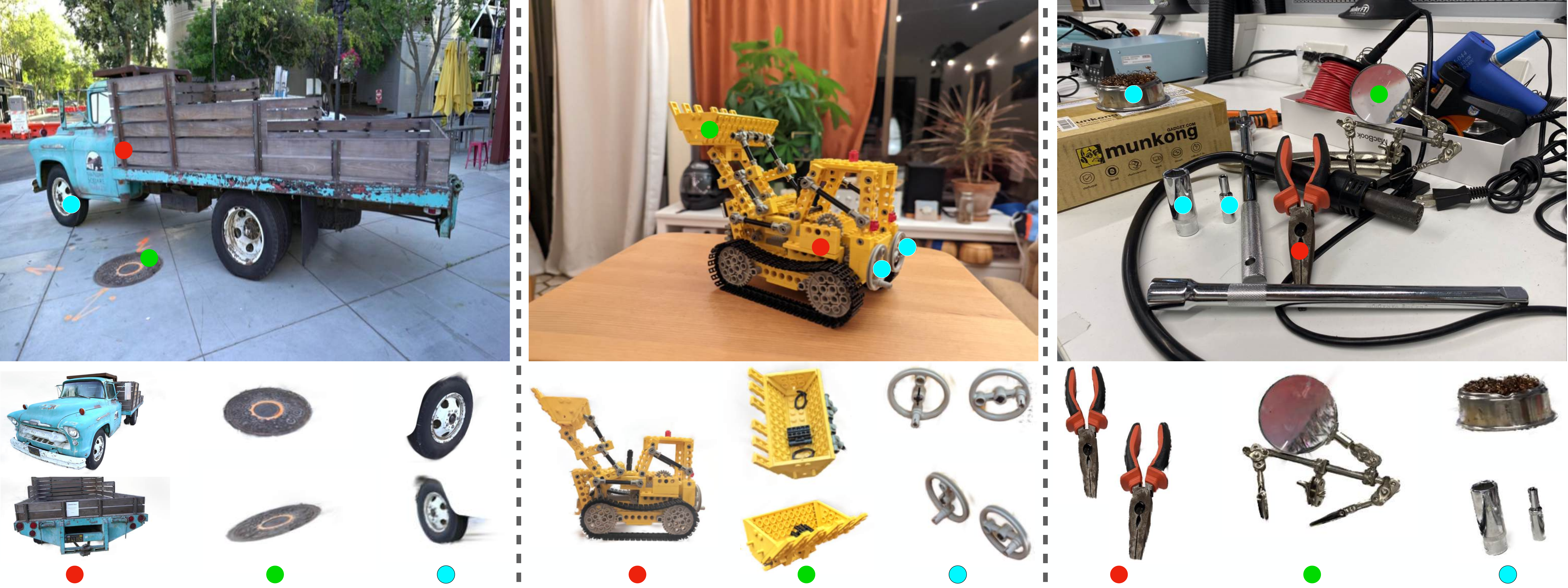}
    \caption{Visualization results of different objects in the following scenes:  truck from Tanks and Temples~\cite{knapitsch2017tanks}, kitchen from Mip-NeRF 360~\cite{barron2022mip}, tools from Shiny~\cite{wizadwongsa2021nex}. }
    \label{fig:quant_paper}
\end{figure}


\textbf{Datasets:} For quantitative evaluation, we test the scenes from LLFF~\cite{mildenhall2019local}, Shiny~\cite{wizadwongsa2021nex}, SPIn-NeRF~\cite{mirzaei2023spin}, and 3D-OVS~\cite{liu2023weakly}. All selected scenes from the LLFF and Shiny datasets are real-world front-facing scenes, with 20-62 images each. SPIn-NeRF provides a collection of scenes from some of the widely-used NeRF datasets~\cite{fridovich2022plenoxels,knapitsch2017tanks,yen2022nerf,mildenhall2019local,mildenhall2021nerf}. It contains a combination of front-facing and \(360^{\circ}\) inward-facing real-world scenes. 3D-OVS contains scenes featuring long-tail objects. 





\textbf{Input types:} Our model accepts all input processed by SAM-Track~\cite{cheng2023segment}. It uses grounding-DINO~\cite{liu2023grounding} to process text inputs. For the LLFF scenes used in NVOS~\cite{ren2022neural}, we follow their input scribbles to obtain the initial mask. For SPIn-NeRF and Shiny, we use clicks (each scene typically requires 1-4 clicks). For the 3D-OVS dataset evaluation, we use text query as input (results in Table~\ref{table:3dovs}). 

\textbf{Evaluation metrics:} Different Image-Based Rendering (IBR) models represent 3D scenes in different ways. Thus, obtaining universal ground-truth 3D masks is difficult. To avoid this challenge, we evaluate the segmentation mask of the projected 2D rendering from the scene. The ground-truth 2D masks are typically obtained from professional image segmentation tools. NVOS provides one ground-truth mask for every scene in LLFF. SPIn-NeRF and 3D-OVS provide masks for multiple images in every scene. Shiny dataset does not contain any ground-truth masks so we create our own ground-truth mask. For evaluation, we generate 2D foreground masks by rendering the Gaussians from the desired viewpoint. We use pixel classification accuracy (Acc)
and foreground intersection-over-union (IoU) for evaluating the segmentation masks.

Following NVOS, we also assess the photo-realistic appearance of the segmented object by rendering it against a black background. We trim both the rendered image and the ground-truth image to the foreground object by applying a bounding box that fits the ground-truth mask. This prevents the evaluation from being biased by the background, especially when the object of interest is relatively small. The metrics we report are  PSNR, SSIM~\cite{wang2004image}, LPIPS~\cite{zhang2018unreasonable}.

\textbf{Implementation details: } To obtain segmentation masks from the user inputs (used in Section~\ref{met:trace}), we leverage the advancements in video segmentation models. The user selects the foreground objects on $\mathbf{I}_0$, and we obtain dense masks for multiple views using SAM-Track~\cite{cheng2023segment}. Note that the use of the video segmentation model is done to enhance the performance further and our method can also work with a single image mask (Table~\ref{table:shiny}). We use KD-Tree for efficiently finding the $k$ nearest neighbors to construct the edges between the nodes.

For all the evaluations, we resize the longer image size to 1008, as commonly practiced in novel-view synthesis. We optimize 3DGS model for each of the considered scenes, without making any changes to the original 3DGS code. For coarse splatting, we keep the cut-off threshold $\tau = 0.9$ for front-facing views and $\tau = 0.3$ for the \(360^{\circ}\) inward-facing scenes. This disparity stems because parts of objects might not be observed from every viewpoint for the latter and also because of the relative ineffectiveness of video tracking for inward-facing scenes (Figure~\ref{fig:truck}). For graph cut, we keep $\gamma = 0.1$ for neighboring pairwise position weights and $\gamma = 1$ for all other weights, the number of neighbors for every node as 10, and the number of clusters for high-confidence nodes as 4 for sink and 1 for source. $\lambda$, $\lambda_n$, $\lambda_u$ can be adjusted depending on the scene and the quality of coarse splatting but generally, $\lambda_n = \lambda_u =  1$ and $\lambda = 0.5$ give decent results.


\textbf{Baselines: } Our comparison includes a selection of baseline models such as NVOS~\cite{ren2022neural}, MVSeg~\cite{mirzaei2023spin}, Interactive segmentation of radiance fields (ISRF)~\cite{goel2023interactive}, Segment Anything in 3D with NeRFs (SA3D)~\cite{cen2024segment}, Segment Any 3D Gaussians (SAGA)~\cite{cen2023segment}, Segment Anything in 3D Gaussians (SAGD)~\cite{hu2024semantic}, Gaussian Grouping~\cite{ye2023gaussian}, and LangSplat~\cite{qin2023langsplat}. Unlike our approach, SAGA, Gaussian Grouping, and LangSplat alter the Gaussian optimization process by learning additional features per Gaussian that increases the optimization time (Table~\ref{table:time}). SAGD is a concurrent work also designed for 3DGS segmentation and has not yet been published. Thus, their results may be subject to change. SAGD, similar to our approach, does not require any segmentation-aware training and uses a cross-view label voting approach to segment selected objects. All the baselines allow for selecting objects using clicks, except LangSplat, for which we use text queries. Further details on baseline implementation are provided in appendix section~\ref{app:baseimp}.
\begin{table}[b]
\parbox{.45\linewidth}{
\centering
\caption{Quantitative results for 2D mask segmentation on NVOS dataset~\cite{ren2022neural}.}
\label{tab:nvos_seg}
\resizebox{0.99\linewidth}{!}{%
\rowcolors{2}{white}{gray!15}
\begin{tabular}{lcc}
\hline
Method & IoU (\%)$\uparrow$ & Acc (\%)$\uparrow$ \\ \hline
graph cut (3D)~\cite{ren2022neural,rother2004grabcut} & 39.4 & 73.6 \\
NVOS~\cite{ren2022neural} & 70.1 & 92.0 \\
ISRF~\cite{goel2023interactive} & 83.8 & 96.4 \\
SA3D~\cite{cen2024segment}  & 90.3 & 98.2 \\
SAGD~\cite{hu2024semantic} & 72.1 & 91.7 \\
SAGA~\cite{cen2023segment} & 90.9 & {98.3} \\
Gaussian Grouping~\cite{ye2023gaussian} & 90.6 & 98.2 \\
LangSplat~\cite{qin2023langsplat} & 74.0 & 94.0\\
GaussianCut (Ours) &  \textbf{92.5} &  \textbf{98.4} \\ \hline
\end{tabular}}
}
\hfill
\parbox{.45\linewidth}{
\centering
\caption{Quantitative results on the SPIn-NeRF dataset~\cite{mirzaei2023spin}.}
\label{table:iou_spin}
\resizebox{0.99\linewidth}{!}{%
\rowcolors{2}{white}{gray!15}
\begin{tabular}{lcc}
\hline
Method   & IoU (\%)         & Acc (\%)   \\ \hline
MVSeg~\cite{mirzaei2023spin}  &  90.4& 98.8          \\
ISRF~\cite{goel2023interactive}   &  71.5 & 95.5     \\
SA3D~\cite{cen2024segment}   &  91.9 & 98.8     \\
SAGD~\cite{hu2024semantic}   &  89.7 & 98.1     \\
SAGA~\cite{cen2023segment}     & 88.0& 98.5    \\
Gaussian Grouping~\cite{ye2023gaussian} & 88.4 & 99.0 \\
LangSplat~\cite{qin2023langsplat} & 69.5 & 94.5  \\
GaussanCut (Ours) &\textbf{ 92.9}   & \textbf{99.2 }     \\
\hline
\end{tabular}}
}
\end{table}
\begin{table}[t]
\parbox{.53\linewidth}{
\centering
\caption{Object rendering results on NVOS~\cite{ren2022neural}.}
\label{tab:psnr_llff}
\resizebox{0.99\linewidth}{!}{%
\rowcolors{2}{white}{gray!15}
\begin{tabular}{lccc}
\hline
Metrics & SSIM$\uparrow$  & PSNR (dB)$\uparrow$  & LPIPS$\downarrow$ \\ \hline
graph cut (3D)~\cite{ren2022neural,rother2004grabcut} & 0.600 & 15.03 & 0.415  \\
NVOS~\cite{ren2022neural} &0.767 &18.40 &0.213 \\
SA3D~\cite{cen2024segment} &  0.794& 20.76& 0.198\\
GaussianCut (Ours) & \textbf{0.840}& \textbf{22.45} & \textbf{0.132} \\ \hline
\end{tabular}}
}
\hfill
\parbox{.43\linewidth}{
\centering
\caption{Quantitative results on Shiny~\cite{wizadwongsa2021nex}.}
\label{table:shiny}
\resizebox{0.99\linewidth}{!}{%
\rowcolors{2}{white}{gray!15}
\begin{tabular}{lcc}
\hline
Scenes   &  IoU (\%)$\uparrow$         & Acc (\%)$\uparrow$            \\ \hline
SA3D~\cite{cen2024segment}    &93.3  &    98.5       \\
SAGD~\cite{hu2024semantic} & 83.3 & 84.7 \\
Coarse Splatting   & 94.3&   99.4    \\
GaussianCut (Ours) &     \textbf{95.0}&      \textbf{99.5} \\ \hline
\end{tabular}}
}
\end{table}
\section{Results}
\subsection{Quantitative results}
\textbf{Dataset from NVOS: } We take the seven scenes from LLFF dataset used in NVOS.  NVOS contains a reference image with input scribbles and a target view with an annotated 2D segmentation mask. As shown in Table~\ref{tab:nvos_seg}, GaussianCut outperforms other approaches. Unlike ISRF, SAGA, Gaussian Grouping, and LangSplat, GaussianCut works on pretrained representations and does not require any changes to the training process. Owing to the fast rasterization, 3DGS-based approaches can also render foreground Gaussians in real-time. To compare the rendering quality of the segmented objects using 3DGS, we train a NeRF model at the same resolution and segment it using SA3D. Table~\ref{tab:psnr_llff} shows the photo-realistic quality of the foreground image against a black background. Gaussian Splatting provides significant gains over NVOS and SA3D for rendering quality, providing a boost of $+4.05$ dB PSNR and $+1.69$ dB PSNR, respectively.




\textbf{Dataset from SPIn-NeRF:} We compare our model on all scenes from the SPIn-NeRF dataset, which includes four $360^\circ$ inward-facing scenes and six front-facing scenes. Our model gives an overall better performance compared to other baselines. Compared to MVSeg, on $360^{\circ}$ scenes such as lego and truck, GaussianCut provides an absolute IoU gain of 14.3\% and 10.5\%, respectively. Our model also performs better compared to other 3DGS baselines as shown in Table~\ref{table:iou_spin}. The $360^{\circ}$ scenes for ISRF were run at one-fourth resolution due to memory constraints.
We show the scene-wise results in Table~\ref{table:iou_spin_per}. Feature-based 3DGS segmentation methods, such as Gaussian Grouping and LangSplat, outperform GaussianCut on certain scenes, but their interactivity can be limited if the optimized features do not 
delineate the object of interest. We show such cases in Figure~\ref{fig:enter-label}. Moreover, we also show in Figure~\ref{fig:spinnerf_mask} that segmentation masks from GaussianCut contain finer details and a better segmentation quality than the ground-truth masks provided in SPIn-NeRF.

\textbf{Dataset from Shiny:}  We test the segmentation performance of our model on four scenes from the Shiny dataset: tools, pasta, seasoning, and giants. 
We create ground-truth masks for 4 test-view images for each scene and compare our model against non-feature learning-based baselines: SA3D and SAGD. We also report the performance of Coarse Splatting (no graph cut) in Table~\ref{table:shiny}. Figure~\ref{fig:shiny} also shows the quality of segmented images for all the scenes.


\subsection{Qualitative results}
\begin{wraptable}{r}{0.5\textwidth}
\centering
\vspace{-15mm}
\caption{Ablation of the energy function averaged over the seven scenes from LLFF dataset.}
\label{table:energy}
\rowcolors{2}{white}{gray!15}
\begin{tabular}{lc}
    \hline
    Energy   & {NVOS} \\ \hline
    Single & 86.6 \\
    Coarse    & 91.2    \\
    w/o \textit{n-link} spatial similarity     &92.4   \\
    w/o \textit{n-link} color similarity   & 92.3     \\
    w/o \textit{t-link} cluster similarity & 91.5    \\
    GaussianCut (Ours) &92.5   \\ \hline
    \end{tabular}
\end{wraptable}
Similar to SA3D, we perform segmentation in three modes: object segmentation, part segmentation, and text-prompting based segmentation. Figure~\ref{fig:quant_paper} shows object and part segmentation. GaussianCut can retrieve complex objects (such as truck, lego bulldozer, mirror) precisely. It can segment smaller part segmentation (manhole cover, lego wheels, and socket wrenches). Our method can also retrieve multiple objects together (socket wrenches and metallic bowl are extracted together). 

\begin{figure}[t]
    \centering
    \includegraphics[width=0.9\linewidth]{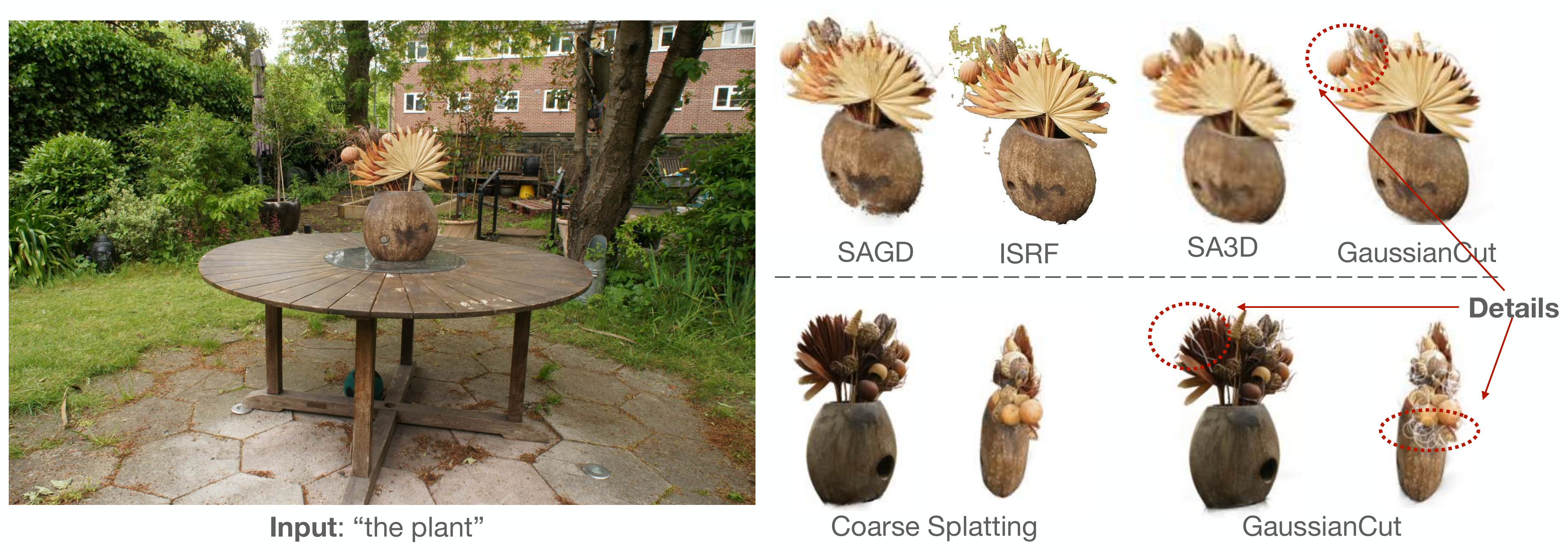}
    \caption{Qualitative comparison: 3D segmentation results of GaussianCut using text on 360-garden~\cite{barron2022mip} scene. Compared to ISRF~\cite{goel2023interactive}, SA3D~\cite{cen2024segment}, SAGD~\cite{hu2024semantic}, GaussianCut segment contain finer details. The graph cut component of GaussianCut also retrieves fine details (like decorations on the plant) that are missed in coarse splatting.}
    \label{fig:bonsai_qual}
\end{figure}

Figure~\ref{fig:bonsai_qual} demonstrates the performance of objection selection using text input. We do a qualitative comparison of GaussianCut with ISRF, SA3D, and SAGD. Feature-based 3DGS methods run into memory issues for this scene. Out of these, SA3D and SAGD also use a text-based prompt to segment the plant. ISRF uses stroke for segmentation. GaussianCut retrieves finer details in the plant with a higher perceptual quality. We also show the rendered image from different viewpoints. It can also be seen that coarse splatting (before the Gaussian graph cut) misses finer details, such as the decorations on the plant, which can be retrieved using GaussianCut.

\subsection{Ablation and sensitivity study}

\textbf{Number of views:} To obtain the 2D segmentation masks, we run the camera on a fixed trajectory and get rendering from different viewpoints (spiral trajectory for front-facing and circular for $360^{\circ}$ inward scene). We limit the number of frames to 30 for front-facing and 40 for $360^{\circ}$ scenes. 
Using more segmentation masks can boost performance, however it might not be always preferred, especially for scenes with a large number of training views. SAM-Track can also handle segmentation for unordered multi-frame images.  Table~\ref{table:views} shows the effect of varying the number of masks on two scenes from the SPIn-NeRF dataset (images used were unordered). Since 3DGS offers fast rasterization, the overall time cost for segmentation does not grow linearly with the number of masks as the time taken for segmentation dominates. We also show the qualitative performance with a single mask (no video segmentation model) and just scribbles (no image segmentation model) in Figure~\ref{fig:ablation_fort}. 

\begin{table}[b]
\centering
\caption{Performance of GaussianCut with varying the number of views passed to the video segmentation models. The number in parenthesis is the percentage of total views for the scene.}
\label{table:views}
\rowcolors{2}{white}{gray!15}
\begin{tabular}{lcccc}
    \toprule
    Number of views   & 5 (10\%) &  9(20\%) & 21 (50\%) & 43 (100 \%) \\ \midrule
    Coarse Splatting on Fortress    &96.1& 96.3  & 96.5 & 96.8  \\
    GaussianCut on Fortress    & 97.7&  97.8  & 97.8 & 97.9\\
    Time Cost (s)  & 51& 55  & 59  & 71\\ \hline \midrule
    Number of views   & 11 (10\%) &  21(20\%) & 51 (50\%) & 102 (100 \%) \\ \midrule
    Coarse Splatting on Lego    &85.5& 88.0 & 88.4 & 88.9   \\
    GaussianCut on Lego    & 87.3&  89.1 &  89.2& 89.2\\
    Time Cost (s)  &  58& 62& 72 & 90  \\ \hline
    \end{tabular}
\end{table}

\textbf{Sensitivity of each term in the graph cut energy function:} In order to understand the contribution of each term in the energy function, Table~\ref{table:energy} shows the average IoU on the NVOS dataset with each term removed from Equation~\ref{eq:nlink} and~\ref{eq:tlink}. Each term contributes to the overall performance and the cluster similarity, in particular, gives a significant boost. 

\textbf{Sensitivity of graph cut hyper-parameters:} 
We test the sensitivity of our Gaussian graph cut algorithm on the number of neighbors (number of edges for each node) and the number of high-confidence clusters. As the number of neighbors increases, the number of edges in the graph also increases (so does the time taken for graph cut). As seen in Table~\ref{tab:ablneigh}, adding more edges can help in modeling more long-distance correlations. However, after a limit, the effects of adding more edges diminish. Adding a large number of clusters for the high-confidence nodes, in Table ~\ref{tab:ablation_clusters}, does not affect the performance drastically and the optimal number can vary depending on the scene. We show sensitivity to other hyper-parameters in appendix section~\ref{sec:appablation}.


\section{Discussion}\label{sec:discussion}

Our results demonstrate that 3DGS allows for direct segmentation using a pretrained model. Developments in 2D segmentation and tracking have played a crucial role in 3D segmentation. We observe that GaussianCut not only generates 3D consistent masks but also improves the segmentation quality of 2D masks by capturing more details (Figure~\ref{fig:2d_masks}). This is more prominent for $360^{\circ}$ scenes, where the tracker can partially or fully miss the object of interest (Figure~\ref{fig:truck}). 
\begin{table}[t]
\parbox{.45\linewidth}{
\centering
\caption{Ablation on the number of neighbors.}
\label{tab:ablneigh}
{%
\rowcolors{2}{white}{gray!15}
\begin{tabular}{lcccc}
\hline
\#Neighbors & 1 & 10 & 50 & 100  \\ \hline
Horns &  91.9 & 93.6 & 93.8 &94.3  \\
Time (s) & 18 & 57 & 209 & 410 \\ \hline
Truck & 93.3 & 95.7  &95.3 &95.2 \\ 
Time (s) & 32 & 96 & 393&  738\\ \hline
\end{tabular}}
}
\hfill
\parbox{.45\linewidth}{
\centering
\caption{Ablation on the number of clusters for high-confidence nodes.}
\label{tab:ablation_clusters}
{%
\rowcolors{2}{white}{gray!15}
\begin{tabular}{lcccc}
\hline
\#Clusters & 1 & 5  & 10 & 20 \\ \hline
Fortress &97.3 &97.8& 97.6 &97.5\\
Horns & 93.8& 93.9 &94.0 & 94.0\\
Truck & 95.6 & 95.7 &95.6  & 95.5\\ 
\hline
\end{tabular}}
}
\end{table}

\textbf{Time requirement:} Since we use pretrained 3DGS models, the optimization time for the Gaussians remains the same as ~\cite{kerbl20233d} (it took under 15 minutes for every scene we use). For inference, masked rasterization of Gaussians is fast and the time taken for graph cut grows roughly linearly with the number of Gaussians. Table~\ref{table:time} shows a detailed breakdown of time taken in each step: preprocessing (obtaining the features from 2D image/video segmentation models), fitting time (3DGS optimization time), and segmentation time (time taken to obtain the segmented output). Compared to feature-based methods, like Gaussian Grouping, LangSplat, and SAGA, our method does not require any alteration to the fitting process and, therefore, has a shorter fitting time. While the segmentation time is higher for GaussianCut, it still has a much shorter overall time. All reported times are on NVIDIA RTX 4090 GPU. 

\begin{table}[h]
\centering
\caption{Comparison of segmentation time (in seconds) on the  NVOS benchmark.}
\label{table:time}
\resizebox{0.9\linewidth}{!}{%
\rowcolors{2}{white}{gray!15}
\begin{tabular}{lcccc}
    \toprule
    Method   & Preprocessing time  &  Fitting time & Segmentation time & Performance (IoU) \\ \midrule
    SAGA~\cite{cen2023segment}  & 71.17 $\pm$ 22.74 & 1448.50 $\pm$ 205.07  & 0.35 $\pm$ 0.05  & 90.9\\ 
    Gaussian Grouping~\cite{ye2023gaussian}    &13.72 $\pm$ 4.63 & 2096.07 $\pm$ 251.96   & 0.55 $\pm$ 0.09 & 90.6  \\
    LangSplat~\cite{qin2023langsplat}    & 2000.34 $\pm$ 1222.19 &  1346.92 $\pm$ 247.00  & 0.82 $\pm$ 0.02 & 74.0\\

    Coarse Splatting (Ours)   & 6.11 $\pm$ 0.38 &  510.97 $\pm$ 106.42 & 19.48 $\pm$ 4.31 & 91.2  \\ 
    GaussianCut (Ours)  & 6.11 $\pm$ 0.38& 510.97 $\pm$ 106.42 & 88.77 $\pm$ 33.68 & 92.5   \\ \hline
    \end{tabular}
}
\end{table}
\textbf{Memory requirement:} While 3DGS has a higher footprint than NeRF-based models, several recent works reduce the memory footprint with limited loss of quality~\cite{niedermayr2023compressed,papantonakis2024reducing,navaneet2023compact3d,fan2023lightgaussian}. Our method only stores one additional parameter $w_g$ for every Gaussian and is less memory-intensive than methods requiring learning a feature field~\cite{felzenszwalb2004efficient,cen2023segment}.

\textbf{Limitations:} 
GaussianCut can address some inaccuracies in 2D video segmentation models, but it may still lead to partial recovery when the initial mask or tracking results are significantly off (Figure~\ref{fig:cycle}). While GaussianCut does not require additional training time, our method can still take up to a few minutes for the graph cut component, which makes the segmentation not real-time. The implementation could be improved by applying graph cut on a subset of Gaussians. We leave this as a future work. Additionally, extending our energy function to include a feature similarity term (in equation~\ref{eq:energy_function}) is another potential improvement. 
We also discuss some failure cases in section~\ref{sec:limitation}. 

\section{Conclusion}
In this paper, we introduce GaussianCut, a novel approach that taps into the underlying explicit representation of 3D Gaussian Splatting to accurately delineate 3D objects. Our approach takes in an optimized 3DGS model along with sparse user inputs on any viewpoint from the scene. We use video segmentation models to propagate the mask along different views and then track the Gaussians that splat to these masks. In order to enhance the precision of partitioning the Gaussians, we model them as nodes in an undirected graph and devise an energy function that can be minimized using graph cut. Our approach shows the utility of explicit representation provided by 3DGS and can also be extended for downstream use cases of 3D editing and scene understanding.

\bibliographystyle{plainnat}
\bibliography{main}


\appendix
\input{appendix}


\clearpage
\end{document}

%% file: appendix.tex








\title{GaussianCut: Interactive segmentation via graph cut for 3D Gaussian Splatting}

%

\author{%
  David S.~Hippocampus\thanks{Use footnote for providing further information
    about author (webpage, alternative address)---\emph{not} for acknowledging
    funding agencies.} \\
  Department of Computer Science\\
  Cranberry-Lemon University\\
  Pittsburgh, PA 15213 \\
  \texttt{hippo@cs.cranberry-lemon.edu} \\
}

\newpage
\section{Implementation details}
We run our segmentation algorithm on 3D Gaussian Splatting representations, following the code provided by Kerbl \textit{et al.}~\cite{kerbl20233d}. All scenes are optimized for 30,000 steps using the default parameters. We use SAM-Track~\cite{cheng2023segment} as the video segmentation model. 

For the datatset used in NVOS~\cite{ren2022neural}, we use the provided reference image with the user scribbles for a fair comparison. For the SPIn-NeRF dataset~\cite{mirzaei2023spin}, we use the first image in the directory as the reference image for the user input. The scenes reported throughout the paper are selected from the following datasets:
\begin{itemize}
\setlength{\itemsep}{0pt} 
\setlength{\parskip}{0pt} 
    \item NVOS (LLFF~\cite{mildenhall2019local} subset): flower, fortress, fern, horns, orchids, trex, leaves
    \item SPIn-NeRF (collection from some widely used datasets~\cite{fridovich2022plenoxels,knapitsch2017tanks,yen2022nerf,mildenhall2019local,mildenhall2021nerf}): orchids, leaves, fortress, horns, truck, lego bulldozer
    \item Shiny~\cite{wizadwongsa2021nex}: giants, tools, seasoning, pasta
    \item Mip-NeRF~\cite{barron2022mip}: cycle, garden, bonsai
    \item LERF~\cite{kerr2023lerf}: figurines 
    \item 3D-OVS~\cite{liu2023weakly}: lawn, sofa, bed, bench, room
\end{itemize}

\paragraph{Mask evaluation: } The Gaussians optimized for 3DGS can have ambiguous structures as they are not geometrically constrained. When partitioning the Gaussians as foreground or background, the boundary Gaussians can appear as having shard-like artifacts (or ``spiky'' Gaussians). Since the goal of this work is to effectively characterize a Gaussian as foreground or background, we render the foreground mask by overriding the colors of background Gaussians. To generate object assets, our algorithm can be combined with Gaussian decomposition based approached~\cite{hu2024semantic}.

\subsection{Baseline implementation 
details}\label{app:baseimp}
\begin{figure}[b]
    \centering
    \includegraphics[width=0.9\linewidth]{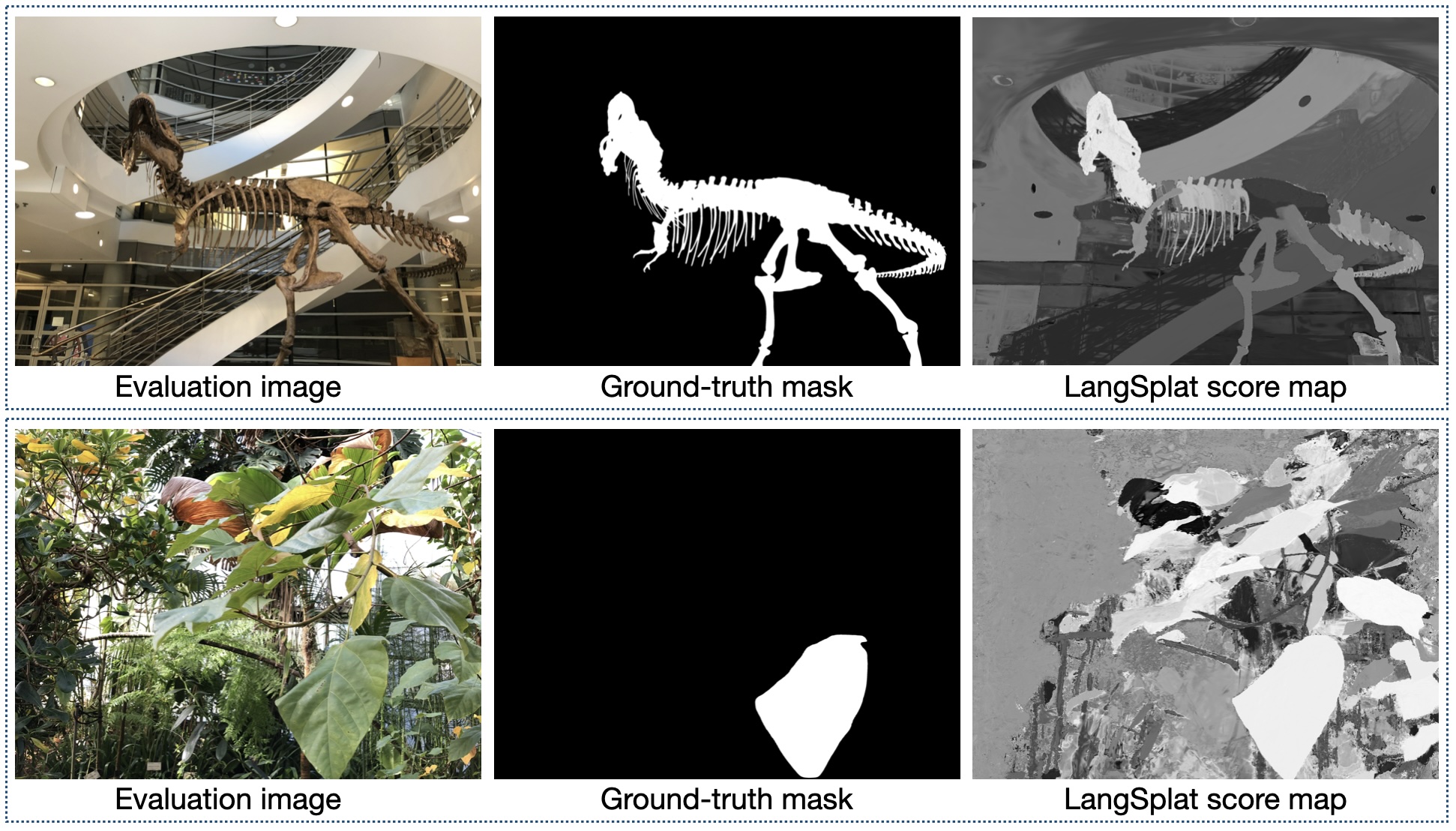}
    \vspace{-3mm}
    \caption{Limitation of LangSplat on Trex and Leaves scenes from NVOS benchmark. Parts of the trex can not be extracted in the top row. In the bottom row, background leaves are also selected along with front leaf.}
    \label{fig:enter-label}
\end{figure}
\subsubsection{Gaussian Grouping}
Gaussian Grouping~\cite{ye2023gaussian} learns an additional feature per Gaussian that can be used to group Gaussians belonging to the same object. We use SAM-Track to get all the 2D segmentation masks. While the default implementation of Gaussian Grouping uses DEVA~\cite{cheng2023segment} masks, we chose SAM-Track for both GaussianCut and Gaussian Grouping to maintain consistency in mask quality across the methods. Similar to GaussianCut, we use clicks to segment objects in NVOS and SPIn-NeRF benchmarks. However, Gaussian Grouping uses a mask for segmenting everything and can, therefore, sometimes produce over-segmented object segments. To handle such cases and obtain the final segmentation, we aggregate all the segments that constitute the object.
We also use the same number of 2D segmented masks as the number of training views. In contrast, for GaussianCut, we limit the number of masks to 30 for front-facing scenes and 40 for $360^{\circ}$ inward-facing scenes. To prevent memory issues in Gaussian Grouping, we restrict the total number of Gaussians across all scenes to $2$M.

\subsubsection{LangSplat}

We obtained all 2D features at feature level ``part'' for LangSplat~\cite{qin2023langsplat}. Since we could not use clicks and scribbles to obtain the segment, we have used text queries. We tried multiple text queries for each scene and reported the results on the best performing query. For certain scenes, text queries can constrain the selection of an object. For instance, in Figure~\ref{fig:enter-label}, multiple instances of the leaves get a high relevancy score when segmenting the front leaf.

\section{Limitations}\label{sec:limitation}
\begin{figure}[ht]
    \centering
    \includegraphics[width=0.9\linewidth]{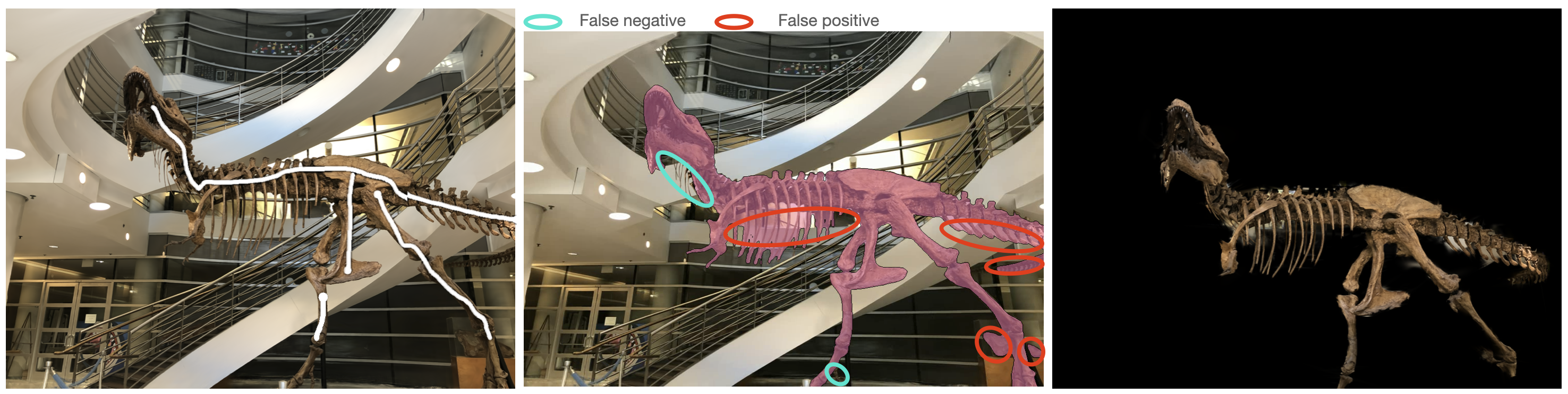}
    \caption{Trex scene from the LLFF~\cite{mildenhall2019local} dataset. \textbf{Left:} reference image with scribbles provided by NVOS~\cite{ren2022neural}. \textbf{Center:} segmentation mask provided by SAM~\cite{kirillov2023segment}. The obtained mask misses finer details and also groups multiple intricate features together. \textbf{Right:} segmented using GaussianCut. While it adds finer details (like near the ribs), the tail still contains some background elements.}
    \label{fig:trex}
\end{figure}
\begin{figure}
    \centering
    \includegraphics[width=0.9\linewidth]{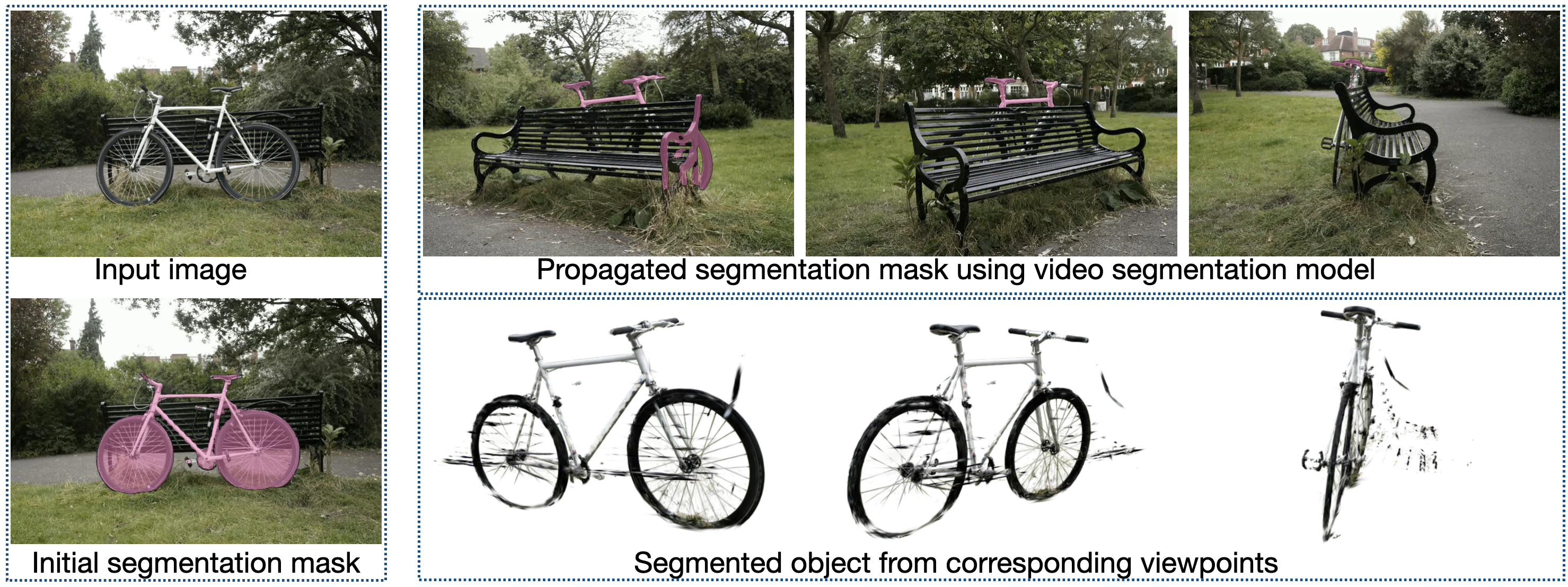}
    \caption{SAM-Track fails to capture major sections of the bicycle when its orientation significantly deviates from the initial position. Even in the reference image, the segmentation mask omits finer details such as the bicycle wheel rims, pedals, and bottle holder. GaussianCut improves segmentation by eliminating substantial portions of the bench to isolate the bicycle, and it partially restores the visibility of the wheel rims.  Despite these improvements, the segmentation remains imprecise.}
    \label{fig:cycle}
\end{figure}
The performance of our method depends on the robustness of image and video segmentation models. For all the scenes tested, we do not tune SAM-Track and use the default settings. SAM-Track (built on SAM) can provide coarse segments, even on the reference image, especially for irregular scenes, as shown in Figure \ref{fig:trex}. GaussianCut improves the segmentation details of SAM but there still remains scope for improving the segmentation performance further for the more intricate patterns.

\begin{figure}
    \centering
    \includegraphics[width=0.9\linewidth]{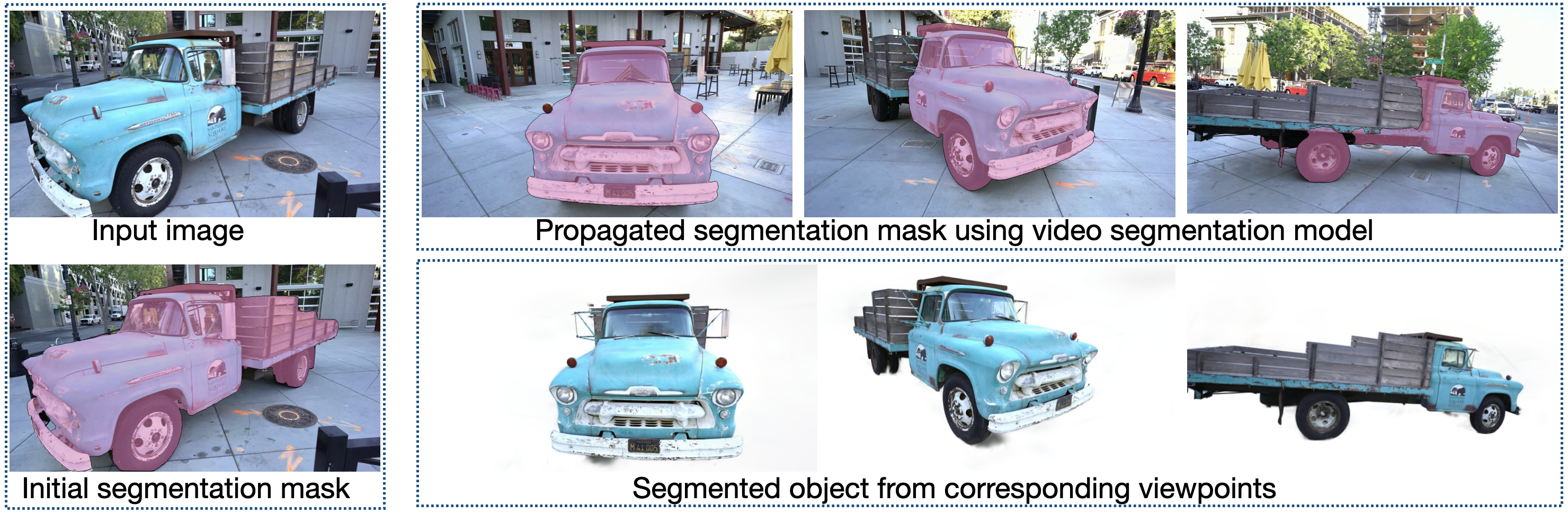}
    \caption{GaussianCut precisely retrieves fine details, such as the mirrors on the front of the truck, even in instances where video-segmentation model struggles to maintain consistency across different views in the scene.}
    \label{fig:truck}
\end{figure}

Similar to image-segmentation models, video-segmentation models can also have inaccurate segmentation masks. This issue is more pronounced in complex 360$^{\circ}$ scenes, where an object can entirely change its orientation, which can lead the trackers to fail in segmenting all views effectively.  We illustrate two instances in Figure \ref{fig:cycle} and \ref{fig:truck}, where GaussianCut corrects the inaccuracies of SAM-Track with varying levels of effectiveness.


\begin{figure}[ht]
    \centering
    \includegraphics[width=0.9\linewidth]{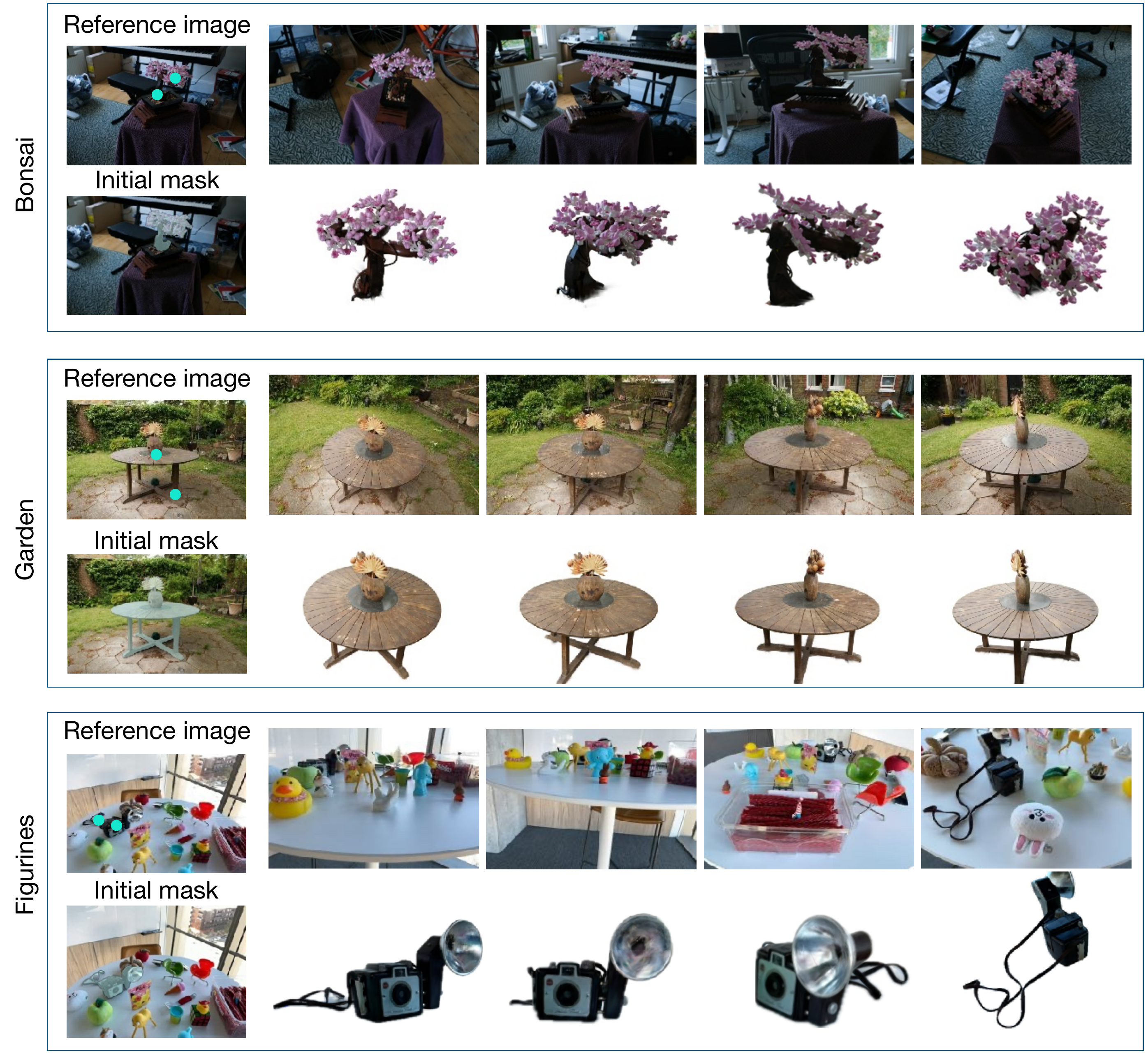}
    \caption{Visualization of selected objects on the Mip-NeRF and LERF dataset. Initial object selection, based on point clicks, and the reference image is shown on the left. }
    \label{fig:quant}
\end{figure}
\section{More visualization results}
We present additional segmentation visualizations for 360$^\circ$ inward scenes taken from Mip-NeRF~\cite{barron2022mip} and LERF datasets~\cite{kerr2023lerf} in Figure \ref{fig:quant}. GaussianCut segments complex and irregular features, including the leaves and wire in the bonsai lego, the detailed decorations of the plant and table in the garden scene, as well as the cord, viewfinder, and flashbulb of the vintage camera.

\section{Additional results}
\subsection{Shiny dataset}
\begin{figure}[ht]
    \centering
    \includegraphics[width=0.8\linewidth]{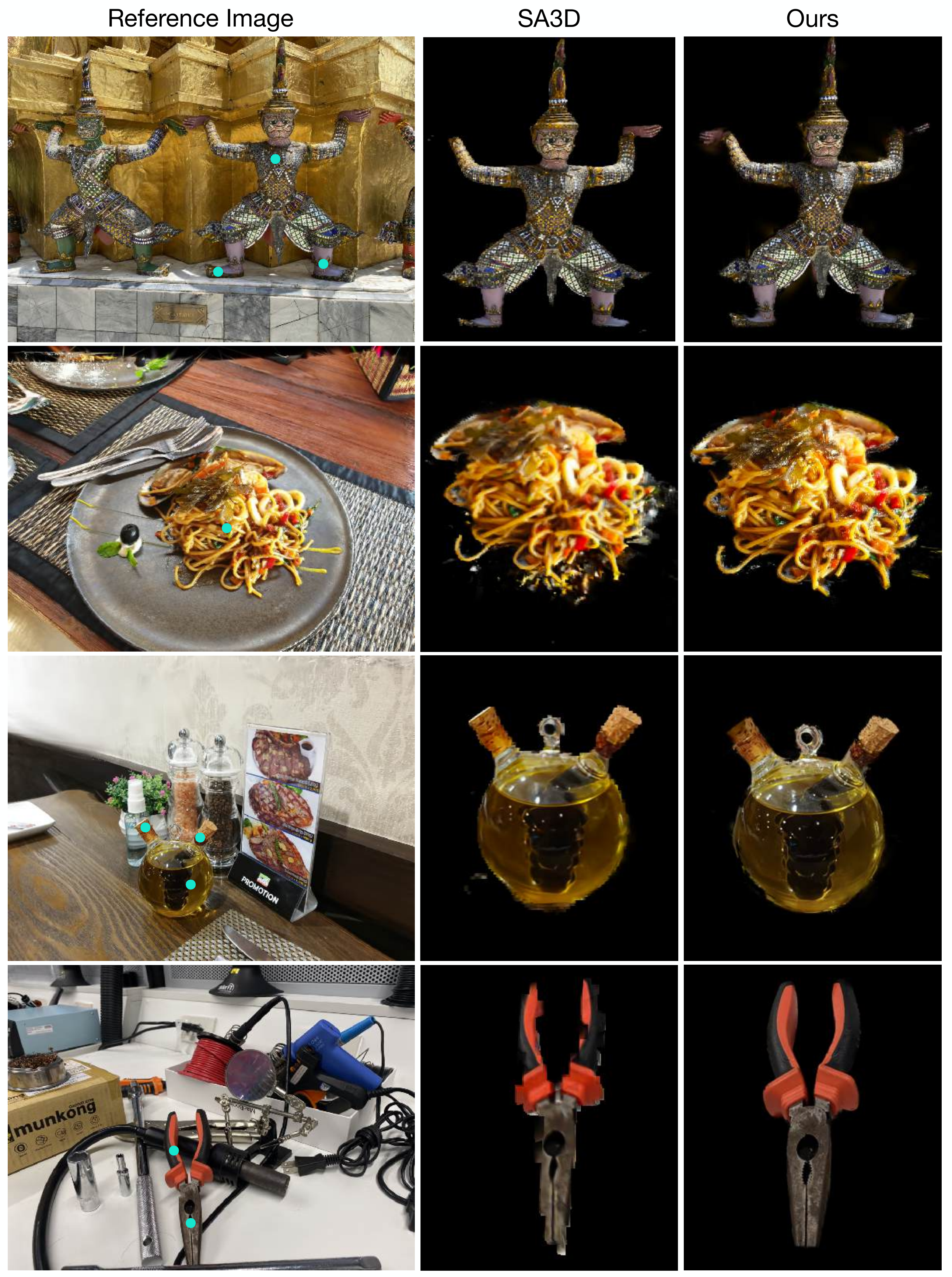}
    \caption{Qualitative results on the Shiny dataset, compared against SA3D~\cite{cen2024segment}. The points used as user inputs are highlighted in the reference image.}
    \label{fig:shiny}
\end{figure}

Scenes in Shiny~\cite{wizadwongsa2021nex} dataset have complex shapes of objects (pasta), are placed in more cluttered environments (tools), or possess a subtler distinction between the foreground and background colors (giants). We test four scenes from the Shiny dataset and label four images from each scene as ground-truth. Table \ref{table:shiny} shows the improvement of GaussianCut against coarse splitting and SA3D~\cite{cen2024segment} with an overall +0.7\%  and +1.7\% absolute gain in foreground mIoU, respectively. Qualitative results from the dataset are shown in Figure \ref{fig:shiny}. GaussianCut can retrieve fine details (like the strands of the pasta) more accurately.

\subsection{SPIn-NeRF}
\begin{figure}[ht]
    \centering
    \includegraphics[width=0.9\linewidth]{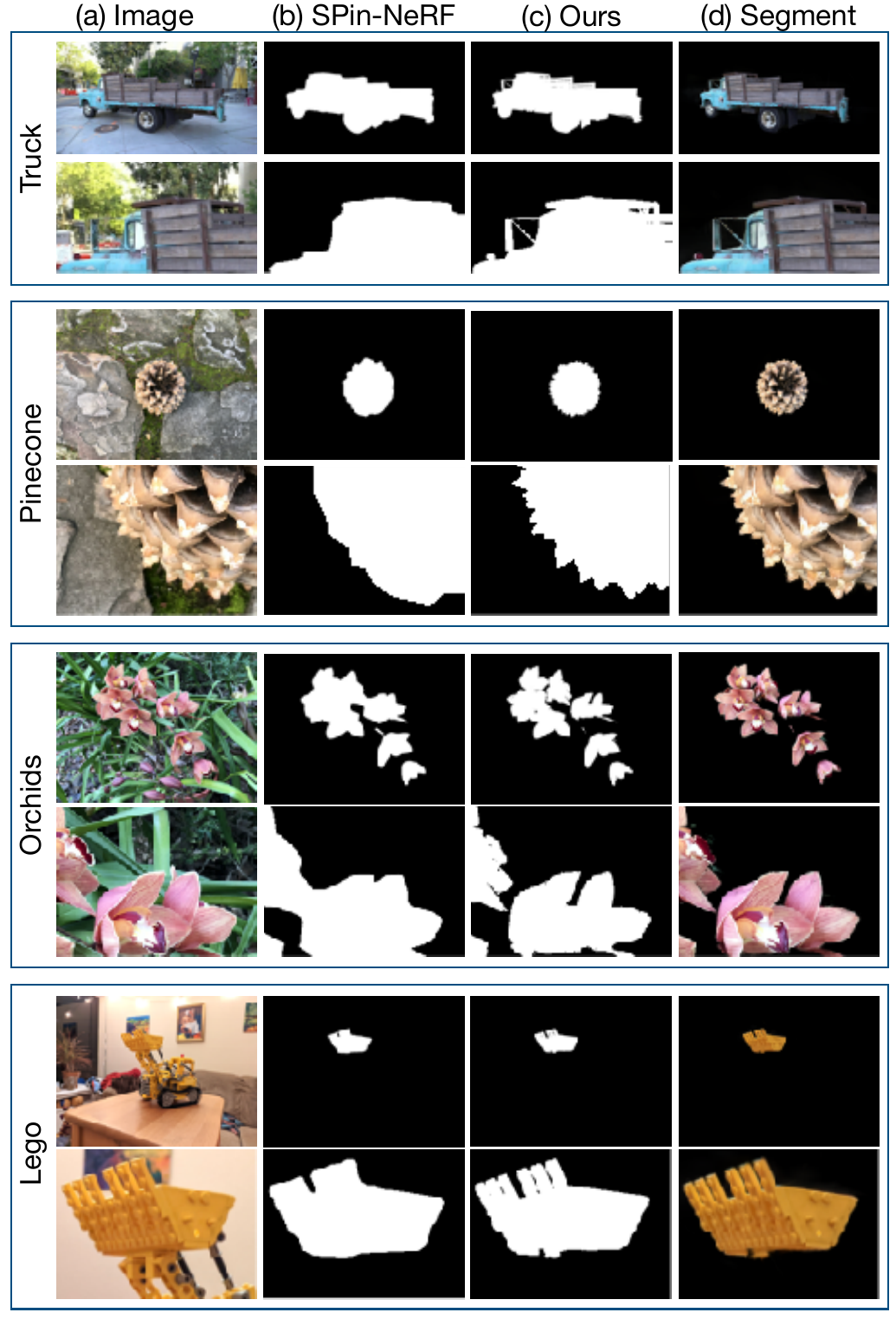}
    \caption{Qualitative comparison of segmentation masks obtained from GaussianCut and the ground-truth used in SPIn-NeRF dataset. }
    \label{fig:spinnerf_mask}
\end{figure}

Table \ref{table:iou_spin_per} shows the performance of GaussianCut on each scene of the SPIn-NeRF dataset. Furthermore, we show in Figure \ref{fig:spinnerf_mask} that the quality of the mask produced by GaussianCut contains finer details than the ground-truth labels from SPIn-NeRF.


\begin{table}
\centering
\caption{Quantitative results on each scene in the SPIn-NeRF dataset.}
\label{table:iou_spin_per}
\resizebox{0.9\linewidth}{!}{%
\rowcolors{2}{white}{gray!15}
\begin{tabular}{lccccccc}
\hline
{Scenes}   & 
{MVSeg~\cite{mirzaei2023spin}} & 
{ISRF~\cite{goel2023interactive}} & {SA3D~\cite{cen2023segment}} & {SAGD~\cite{hu2024semantic}} & 
{Gaussian Grouping~\cite{ye2023gaussian}} & 
{LangSplat~\cite{qin2023langsplat}} & {GaussianCut)} \\ \midrule
Orchids  &  92.7& 84.2 & 83.6 &   85.4 & 89.4  & 88.2 & 87.8\\
Leaves   & 94.9   &  87.2 & 97.2 &  87.9 & 96.7 & 46.5 & 96.3\\
Fern     &   94.3 & 83.1  & 97.1 & 92.0 & 97.3 & 97.3 & 96.1\\
Room     &  95.6  & 67.6  &  88.2  &  86.5 & 96.2 & 64.5 & 95.2\\
Horns    & 92.8  & 84.3 & 94.5 & 91.1 & 92.5 & 87.1 & 93.6\\
Fortress & 97.7    & 92.6 & 98.3 &   96.6 & 97.8 & 95.8 & 97.7\\
\midrule
Fork     & 87.9    & 19.9  & 89.4     &  83.4 & 90.0 & 70.1 & 85.4\\
Pinecone     &  93.4    & 60.0  & 92.9 & 92.0 & 92.2 & 54.4 & 91.9\\
Truck    & 85.2 &79.2 & 90.8 &   93.0 & 95.9 & 53.8 & 95.7\\
Lego     & 74.9    & 57.3 & 92.2   & 88.4  & 36.0 & 36.3 & 89.2\\

\midrule
mean     &  90.9  & 71.5 & 92.4   & 89.7   &  88.4 & 69.5 & 92.9\\ \hline
\end{tabular}
}
\end{table}

\subsection{3D-OVS}
\begin{table}[t]
    \centering
    \caption{Quantitative evaluation on 3D-OVS~\cite{liu2023weakly} dataset. CGC refers to  Contrastive Gaussian Clustering method.}
    \label{table:3dovs}
    \resizebox{0.9\linewidth}{!}{%
    \begin{NiceTabular}{@{}cccccc@{}}
    \CodeBefore
 \rowlistcolors{2}{white,gray!15}[restart,cols={2-6}]
\Body
        \toprule
          &  Item  & Gaussian Grouping~\cite{ye2023gaussian} &  {LangSplat~\cite{qin2023langsplat}} & {CGC~\cite{silva2024contrastive}}   &{GaussianCut (Ours)} \\
         \midrule

        \multirow{7}{*}{\rotatebox[origin=c]{90}{Bed}}& Banana & \underline{96.9} & 17.8 & 95.9 &  \textbf{97.2 }\\
        & Leather Shoe & \underline{97.6} & 71.4 & \textbf{97.9} &  96.4\\
        & Camera & \textbf{95.3} & 3.3 & \underline{85.0} &  \textbf{95.3} \\
        & Hand & \textbf{96.8} & 6.9 & \underline{95.5} &  94.0\\
        & Red Bag & \underline{98.5} & 81.2 & 98.1 &  \textbf{98.8}\\
        & White Sheet & \textbf{99.0}& 25.5 & \textbf{99.0} & \underline{98.9} \\
        \midrule
        & Average & \textbf{97.3} & 34.3 & 95.2 & \underline{96.8}\\ \midrule
        \multirow{8}{*}{\rotatebox[origin=c]{90}{Bench}}& Wall & \textbf{98.7} & 42.6 & \textbf{98.7} &   \underline{97.7}\\
        & Wood & \underline{97.8} & 88.0 & \textbf{98.0} &  96.2\\
        & Egg Tart & \underline{95.4} &  87.4&  \textbf{97.2}&  93.2\\
        & Orange Cat & 94.5 & 96.4 & \underline{97.3} & \textbf{97.6} \\
        & Green grape & 0.0 & 93.7 & 95.5 & 94.6 \\
        & Offroad car & \underline{93.7} & \underline{93.7} & 93.6 & \textbf{95.2} \\
        & Doll & 36.0 & 92.1 & \underline{92.2} & \textbf{93.2} \\ \midrule
        & Average & 73.7 & 84.8 & \textbf{96.1} &  \underline{95.4}\\ \midrule
    \multirow{7}{*}{\rotatebox[origin=c]{90}{Room}}& Wall & \textbf{99.4}&  24.3&  43.6&  \underline{99.1}\\
        &  Chicken& 0.0 & \textbf{94.7} &\underline{91.4}  & 91.2 \\
        & Basket & 87.9 & \textbf{98.3} & \underline{97.4} & 96.3 \\
        & Rabbit & \underline{96.3} & 45.3 & \textbf{97.0} & 93.8 \\
        & Dinosaur & 92.7 & 6.8 & \textbf{94.9} &  \underline{93.1}\\
        & Baseball & \textbf{97.5} & 68.4 & \underline{97.4} & 95.6 \\ \midrule
        & Average &79.0  & 56.3 & \underline{86.9} &  \textbf{94.9}\\ \midrule
      \multirow{8}{*}{\rotatebox[origin=c]{90}{Sofa}}& Pikachu & 48.8& \underline{89.6} & 0.0 & \textbf{94.3} \\
        & UNO cards & \textbf{95.8} & 79.6 & 95.4 & \underline{95.5} \\
        & Nintendo switch & 90.9 & 89.8 & 92.6 & 93.7 \\
        & Gundam & 16.1 & \underline{79.5} & 76.9 &  \textbf{91.0}\\
        & Xbox controller & 59.9 & 67.8 & \underline{96.1} & \textbf{97.7} \\
        & Sofa & \underline{97.4} & 0.0 & 44.0 & \textbf{98.0} \\ \midrule
        & Average & \underline{68.1} & 67.7 & 67.5 & \textbf{95.0} \\ \midrule
         \multirow{7}{*}{\rotatebox[origin=c]{90}{Lawn}} & Apple & \textbf{96.3} & \underline{94.0} & 93.8 & 88.8 \\
        & Cap  & \textbf{98.4} &\textbf{ 98.4 }& \underline{97.9} & 92.0 \\
        & Stapler  & 94.7 & \textbf{96.2} & \underline{95.6} & 88.1 \\
        & Headphone & \textbf{94.5 }& \underline{91.7} & 70.2 & 84.2 \\
        & Hand soap & \textbf{96.0} & \underline{95.2} & 93.7 &91.3  \\ 
        & Lawn & 99.2 & \textbf{99.5} & \underline{99.3} &  94.1\\ \midrule
        & Average &\textbf{ 96.5} & \underline{95.8} & 91.8 &  89.8\\ \bottomrule
        
    \end{NiceTabular}}
\end{table}
To test the performance of our method using text queries, we test on the 3D-OVS~\cite{liu2023weakly} dataset and compare it against Gaussian Grouping~\cite{ye2023gaussian}, LangSplat~\cite{qin2023langsplat}, and Contrastive Gaussian Clustering~\cite{silva2024contrastive} in Table \ref{table:3dovs}. We use the grounding-DINO integration in SAM-Track to obtain the initial segments. The baseline numbers reported in Table \ref{table:3dovs} are taken from ~\cite{silva2024contrastive}. We use a different text query for some objects than ~\cite{silva2024contrastive}. This was done to ensure that we have a decent initial mask from SAM-Track as the goal of our work is not to improve language understanding in 3D models.

\subsection{Qualitative comparison with 2D segmentation model}
\begin{figure}
    \centering
    \includegraphics[width=0.9\linewidth]{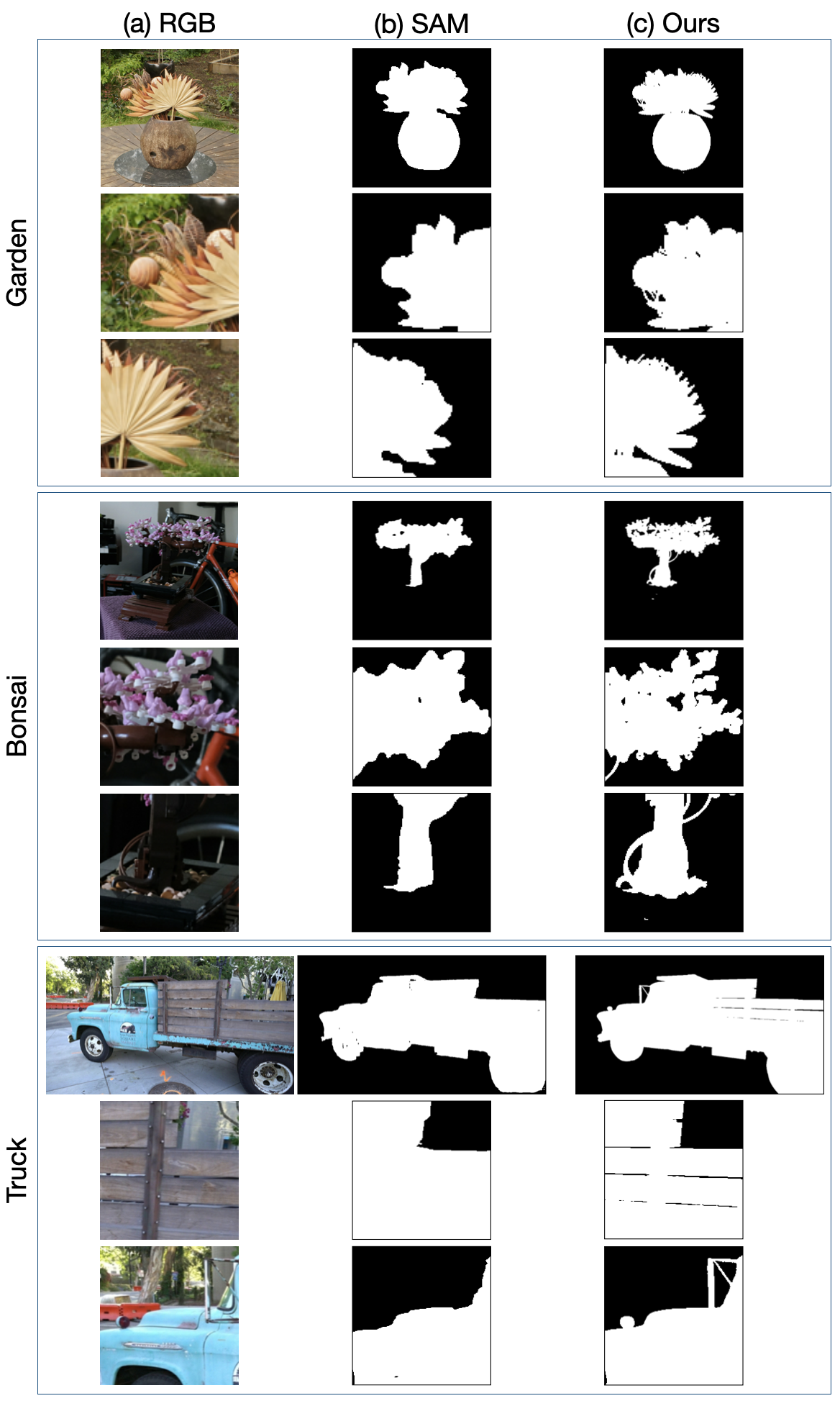}
    \caption{Visualization of segmentation masks from SAM and GaussianCut. }
    \label{fig:2d_masks}
\end{figure}

The objects segmented by GaussianCut exhibit fine details, as depicted in the masks presented in Figure \ref{fig:2d_masks}. Although our method uses SAM predictions as an initial mask, the segregation of Gaussians provides information with greater precision compared to SAM alone.

\section{Additional sensitivity analysis and ablations}\label{sec:appablation}
\subsection{Binary weights for coarse splatting}
As mentioned in Section~\ref{met:trace}, the likelihood term  $w_g$ for each Gaussian $g$ is obtained by taking a weighted ratio of $g$'s contribution on the masked pixels compared to the total number of pixels it affects. Instead of using weighted assignment, we can also have a hard binary assignment where a $g$ either contributes to a foreground pixel or it doesn't. For the $n$ viewpoints, $\mathcal{I} := \{\mathbf{I}^j \}_{j=1}^n$ that have corresponding masks $\mathcal{M} := \{\mathbf{M}^j \}_{j=1}^n$, 
\begin{align}
w_g = \frac{\sum_{j}\sum_{\textbf{p} \in \textbf{M}^j } \mathbb{I}(T_g^j > 0)  }{\sum_j\sum_{\textbf{p} \in {\textbf{I}}^j} \mathbb{I}(T_g^j > 0)},
\end{align}
which reflects the ratio of the number of pixels that $g$ has contributed to in $\mathbf{M}^j$ and $\mathbf{I}^j$.
\begin{table}[t]
\centering
\caption{Comparison of soft and hard weight assignment of $w_g$.}
\label{table:binaryweights}
\rowcolors{2}{white}{gray!15}
\begin{tabular}{lcc}
    \toprule
    Scene   & Soft assignment   &  Hard assignment\\ \midrule
    Fern (NVOS)    &83.06 & 82.56   \\
    Fortress (NVOS)    & 97.97 &  98.12  \\
    Leaves (NVOS)  & 95.95  & 95.60 \\ 
    Lego (SPIn-NeRF)   & 89.18 & 88.95\\
    Pinecone (SPIn-NeRF)   & 91.89 & 91.99\\ \hline
    \end{tabular}
\end{table}
As shown in Table \ref{table:binaryweights}, since soft assignment has marginally better performance, it is our default implementation.

\subsection{Sensitivity of hyperparameters}
We share a default setting in section~\ref{sec:setup} which performs reasonably well on all our datasets. The sensitivity of each parameter can be very scene-dependent. For instance, in a scene where parts of an object have different colors, a very high weight on the color similarity can affect adversely. We show the effect of 
 $\lambda$ (controls the pull of neighboring vs terminal edges) and $\gamma$
 (decay constant of the similarity function) on two scenes in Table~\ref{table:lambda_values} and Table~\ref{table:sigma_values}, respectively. The reported metric is IoU.

\begin{table}[t]
\parbox{.45\linewidth}{
\centering
\caption{Performance comparison for different $\lambda$ values.}
\label{table:lambda_values}
\rowcolors{2}{white}{gray!15}
\begin{tabular}{lcccc}
    \toprule
    Scene   & $\lambda = 0.5$ & $\lambda = 1$ & $\lambda = 2$ & $\lambda = 4$ \\ \midrule
    Fortress & 97.67 & 97.99 & 97.95 & 97.80 \\
    Lego     & 89.15 & 89.18 & 89.18 & 88.49 \\ \bottomrule
\end{tabular}
}
\hfill
\parbox{.45\linewidth}{
\centering
\caption{Performance comparison for different $\gamma$ values.}
\label{table:sigma_values}
\rowcolors{2}{white}{gray!15}
\begin{tabular}{lcccc}
    \toprule
    Scene   & $\gamma = 0.5$ & $\gamma = 1$ & $\gamma = 2$ & $\gamma = 4$ \\ \midrule
    Fortress & 96.12 & 97.95 & 97.56 & 96.04 \\
    Lego     & 89.20 & 89.18 & 89.18 & 89.19 \\ \bottomrule
\end{tabular}
}
\end{table}

\subsection{Threshold of coarse splatting}
For the four 360-degree inward scenes in the SPIn-NeRF benchmark, we show a sweep of the threshold $\tau$ (default is 0.3 in our implementation) used for Coarse Splatting. GaussianCut outperforms all the thresholds considered for coarse splatting as shown in Table~\ref{table:coarse_splatting}. 
\begin{table}[t]
\centering
\caption{Coarse splatting baseline with different thresholds.}
\label{table:coarse_splatting}
\rowcolors{2}{white}{gray!15}
\begin{tabular}{lcc}
    \toprule
    Threshold   & IoU & Acc \\ \midrule
    Coarse@0.1  & $88.47 \pm 4.85$ & $98.96 \pm 0.53$ \\
    Coarse@0.3  & $89.67 \pm 3.18$ & $98.94 \pm 0.72$ \\
    Coarse@0.5  & $87.76 \pm 3.06$ & $98.45 \pm 1.50$ \\
    Coarse@0.7  & $83.30 \pm 6.04$ & $97.58 \pm 2.84$ \\
    Coarse@0.9  & $72.13 \pm 11.26$ & $96.08 \pm 4.69$ \\
    GaussianCut & $90.55 \pm 3.76$ & $99.18 \pm 0.41$ \\ \bottomrule
\end{tabular}
\end{table}

\subsection{How important are the 2D segmentation masks? } 
\begin{figure}
    \centering
    \includegraphics[width=0.6\linewidth]{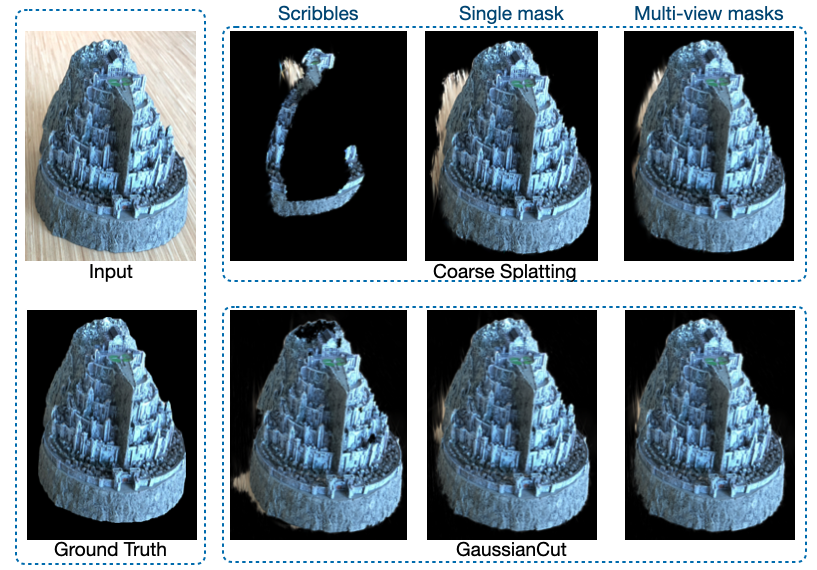}
    \caption{ 
    We compare coarse splatting (w/o graph cut) and GaussianCut. Scribbles refer to using direct input, single mask refers to taking the mask from one viewpoint, and multi-view masks refer to using video segmentation. The effectiveness of GaussianCut becomes more prominent when the inputs are sparse.}
    \label{fig:ablation_fort}
\end{figure}

In order to understand the extent to which the performance of our model depends on the initial 2D segmentation mask, we do the masked rasterization with just scribbles, a single mask, and multi-view masks. Figure \ref{fig:ablation_fort} shows the segmentation result of Coarse Splatting and GaussianCut. The effectiveness of GaussianCut is heightened further when the initial segmentation mask is sparse. Table~\ref{table:segmentation_comparison} also shows the performance improvement when running GaussianCut directly on user scribbles.

\begin{table}
\centering
\caption{Segmentation performance with just user scribbles for NVOS scenes.}
\label{table:segmentation_comparison}
\rowcolors{2}{white}{gray!15}
\begin{tabular}{lccc}
    \toprule
    Scene      & Scribbles & Scribbles (with graphcut) & GaussianCut \\ \midrule
    Fern       & 8.17  & 47.97 & 83.06 \\
    Flower     & 7.48  & 85.30 & 95.37 \\
    Fortress   & 15.12 & 95.67 & 97.95 \\
    Trex       & 6.74  & 50.44 & 83.43 \\
    Orchids    & 6.17  & 85.25 & 95.80 \\
    \bottomrule
\end{tabular}
\end{table}

